\documentclass{article}

\usepackage[nonatbib, final]{neurips_2021}
\usepackage[numbers]{natbib}

\usepackage[utf8]{inputenc} 
\usepackage[T1]{fontenc}    
\usepackage{hyperref}       
\hypersetup{citecolor=MidnightBlue}
\usepackage{url}            
\usepackage{booktabs}       
\usepackage{amsfonts}       
\usepackage{nicefrac}       
\usepackage{microtype}      
\usepackage{xcolor}         
\usepackage{enumitem}

\usepackage{graphicx}
\graphicspath{{./figures/}}
\usepackage[font=small,skip=0pt]{subcaption} 
\usepackage{amsmath,soul,amssymb}
\usepackage{balance}
\usepackage{wrapfig}
\usepackage{dblfloatfix} 
\usepackage{multirow}
\usepackage{bm}
\usepackage{bbm} 
\usepackage{accents} 
\usepackage{fancyhdr} 
\usepackage{listings}  
\usepackage{todonotes}
\usepackage{algorithm}
\usepackage{algorithmic}

\DeclareMathOperator{\R}{\mathbb{R}} 
\DeclareMathOperator{\Y}{\mathcal{Y}} 
\DeclareMathOperator{\lfs}{\bm{\lambda}}  

\newcommand{\softmax}{\mathrm{softmax}}

\newcommand{\weasel}{\texttt{WeaSEL}}
\newcommand{\brackets}[1]{\left( #1 \right)}

\newcommand{\indicator}[1]{\mathbbm{1}\{ #1 \}}
\newcommand{\encoder}{$e $}

\newcommand{\kthLfs}{\lfs^{(k)}}
\newcommand{\kthScore}{\mathbf{s}^{(k)}}
\newcommand{\kthFeatures}{\mathbf{x}^{(k)}}
\newcommand{\features}{\mathbf{x}}

\newcommand{\lambdaOnehot}{\accentset{\rule{.4em}{.8pt}}{\lfs}}

\newcommand{\first}[1]{{\color{blue}\textbf{#1}}}
\newcommand{\second}[1]{{\textbf{#1}}}

\newcommand{\Abl}[2]{{#1 $\pm$ #2}}
\newcommand{\emphAbl}[2]{{\textbf{#1 $\pm$ #2}}}
\newcommand{\emphBadAbl}[2]{{\color{red}\textbf{#1 $\pm$ #2}}}

\title{End-to-End Weak Supervision}

\author{%
   Salva Rühling Cachay\textsuperscript{1,2}\thanks{\href{mailto:salvaruehling@gmail.com}{salvaruehling@gmail.com}.}
  \enskip\enskip\enskip\enskip\enskip\enskip\enskip
   Benedikt Boecking\textsuperscript{1}
  \enskip\enskip\enskip\enskip\enskip\enskip\enskip
   Artur Dubrawski\textsuperscript{1}\\
   \\
   \textsuperscript{1} Carnegie Mellon University 
  \enskip\enskip\enskip\enskip\enskip 
   \textsuperscript{2} Technical University of Darmstadt\\
}

\begin{document}

\maketitle

\begin{abstract}
Aggregating multiple sources of weak supervision (WS) can ease the data-labeling bottleneck prevalent in many machine learning applications, by replacing the tedious manual collection of ground truth labels. 
Current state of the art approaches that do not use any labeled training data, however, require two separate modeling steps: Learning a probabilistic latent variable model based on the WS sources -- making assumptions that rarely hold in practice -- followed by downstream model training. 
Importantly, the first step of modeling does not consider the performance of the downstream model.
To address these caveats we propose an end-to-end approach for directly learning the downstream model by maximizing its agreement with probabilistic labels generated by reparameterizing prior probabilistic posteriors with a neural network. 
Our results show improved performance over prior work in terms of end model performance on downstream test sets, as well as in terms of improved robustness to dependencies among weak supervision sources.

\end{abstract}

\section{Introduction}
\label{introduction}
The success of supervised machine learning methods relies on the availability of large amounts of labeled data. The common process of manual data annotation by humans, especially when domain experts need to be involved, is expensive, both in terms of effort and cost, and as such presents a major bottleneck for deploying supervised learning methods to new domains and applications. 

Recently, data programming, a paradigm that makes use of multiple sources of noisy labels, has emerged as a promising alternative to manual data annotation~\cite{DP}.
It encompasses previous paradigms such as distant supervision from external knowledge bases \cite{distantSupervision2, distantSupervision1}, crowdsourcing \cite{SkeneModel, Crowdsourcing1, Crowdsourcing3, Crowdsourcing2}, and general heuristic and rule-based labeling of data \cite{patternbasedWS, rulebasedWSforChemistry}.
In the data programming framework, users encode domain knowledge into so called labeling functions (LFs), which are functions (e.g. domain heuristics or knowledge base derived rules) that noisily label subsets of data.
The main task for learning from multiple sources of weak supervision is then to recover the sources' accuracies in order to estimate the latent true label, without access to ground truth data. In previous work \cite{DP, Multitask, triplets}, this is achieved by first learning a generative probabilistic graphical model (PGM) over the weak supervision sources and the latent true label to estimate \emph{probabilistic labels}, which are then used in the second step to train a \emph{downstream model} via a noise-aware loss function.

Data programming  has led to a wide variety of success stories in domains such as healthcare \cite{fries2019weakly,dunnmon2020cross} and e-commerce~\cite{Drybell}, but the existing PGM based frameworks still come with a number of drawbacks. 
The separate PGM does not take the predictions of the downstream model into account, and indeed this model is trained independently of the PGM.  
In addition, current approaches for estimating the unknown class label via a PGM need to rely on computationally expensive approximate sampling methods~\cite{DP}, estimation of the full inverse of the LFs covariance matrix \cite{Multitask}, or they need to make strong independence assumptions~\cite{triplets}.  
Furthermore, existing prior work and the associated theoretical analyses make assumptions that may not hold in practice~\cite{DP, Multitask, triplets}, such as availability of a  well-specified generative model structure (i.e.\ that the dependencies and correlations between the weak sources have been correctly specified by the user), that LF errors are randomly distributed across samples, and that the latent label is independent of the features given the weak labels (i.e.\ only the joint distribution between the sources and labels needs to be modeled).
\begin{figure*}
    \centering
    \includegraphics[width=.75\textwidth]{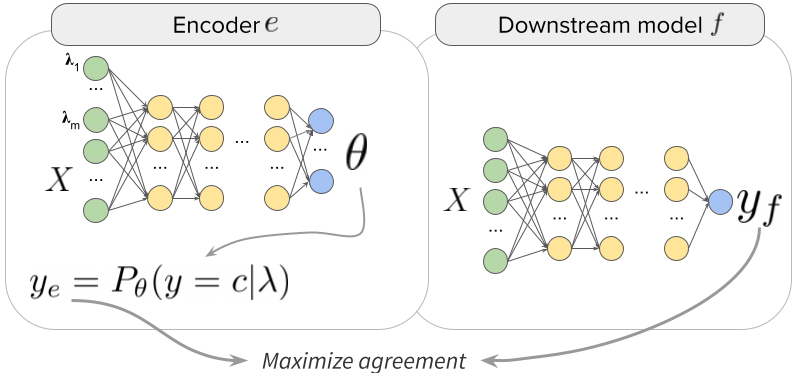}
    \vspace{-1mm}
    \caption{For a task with unobserved ground truth labels $y$, given $m$ sources of weak supervision $\lambda_i$ and training features $X$, \weasel\ trains a downstream model $f$  by maximizing the agreement of its predictions $y_f$ with probabilistic labels $y_e=P_\theta(y=c|\lfs)$ generated by reparameterizing the posterior of prior work with sample-dependent accuracy scores $\theta$ produced by an encoder network $e$.
    }
    \label{fig:diagram}
\end{figure*}

We introduce \weasel, our Weakly Supervised End-to-end Learner model for training neural networks with, exclusively, multiple sources of weak supervision as noisy signals for the latent labels. \weasel\ is based on 1) \emph{reparameterizing previous PGM based posteriors} with a neural encoder network that produces \emph{accuracy scores for each weak supervision source}; and 2) training the encoder and downstream model on \emph{the same target loss, using the other model's predictions as constant targets}, to maximize the agreement between both models.
The proposed method needs no labeled training data, and neither assumes sample-independent source accuracies nor redundant features for latent label modeling. We show empirically that it is not susceptible to highly correlated LFs.
In addition, the proposed approach can learn from multiple probabilistic sources of weak supervision.

Our contributions include:
\begin{itemize}[leftmargin=*]
    \item We introduce a flexible, end-to-end method for learning models from multiple sources of weak supervision. 
    \item We empirically demonstrate that the method is naturally robust to adversarial sources as well as highly correlated weak supervision sources. 
    \item
    We release an open-source, end-to-end system for arbitrary PyTorch downstream models that will allow practitioners to take advantage of our approach\footnote{\url{https://github.com/autonlab/weasel}}.
    \item
    We show that our method outperforms, by as much as 6.1 F1 points, state-of-the-art latent label modeling approaches on 4 out of 5 relevant benchmark datasets, and achieves state-of-the-art performance on a crowdsourcing dataset against methods specifically designed for this setting.
\end{itemize}

\section{Related Work}
\label{RelatedWork}
\textbf{Multi-source Weak Supervision} $\;$
The data programming paradigm~\cite{DP} allows users to programmatically label data through multiple noisy sources of labels, by treating the true label as a latent variable of a generative PGM.
Several approaches for learning the parameters of the generative model have been introduced ~\cite{Multitask, triplets, TripletsMean} to address computational complexity issues. 
Existing methods are susceptible to misspecification of the dependencies and correlations between the LFs, which can lead to substantial losses in performance \cite{MisspecificationInDP}. Indeed, it is common practice to assume a conditionally independent model -- without any dependencies between the sources -- in popular libraries~\cite{Drybell, Snorkel} and related research \cite{SkeneModel, anandkumar2014tensor, Snuba, IWS}, even though methods to learn the intra-LF structure have been proposed \cite{structureLearning1, structureLearning2, SL-static-analysis}. In natural language processing,  \cite{wang2018deep} consider more relaxed and broader forms of weak supervision and introduce a framework that uses virtual evidence as prior belief over latent
labels and their interdependencies, learning an end model jointly with the label model via variational EM. 
As in the approach proposed in this paper, the aforementioned methods do not assume any labeled training data, i.e. the downstream model is learned based solely on outputs of multiple LFs on unlabeled data.
The traditional co-training paradigm \cite{co_training} on the other hand is similar in spirit but requires some labeled data to be available. Recent methods that study the co-training setup where labeled training data supplements multiple WS sources, include \cite{Awasthi2020Learning,astra}.
Note that the experiments in \cite{Awasthi2020Learning, astra} rely on large pre-trained language models, making the applicability of the approach without such models or to non-text domains unclear.
\textbf{Crowdsourcing} $\;$
Aggregating multiple noisy labels is also a core problem studied in the crowdsourcing literature. Common approaches model worker performance and the unknown label jointly~\cite{SkeneModel,Crowdsourcing3,Crowdsourcing2} using expectation maximization (EM) or similar approaches. Some core differences to learning from weak supervision sources are that errors by crowdworkers  are usually assumed to be random, and that task assignment is not always fixed but can be optimized for. 
The benefits of jointly optimizing the downstream model and the aggregator of the weak sources have been recognized in multiple end-to-end methods that have been proposed for the crowdsourcing problem setting~\cite{raykar10a, doctorNet, Crowdsourcing4, khetan2017learning, rodrigues2018deep, MaxMIG}. They often focus on image labeling and EM-like algorithms for modeling and aggregating the workers.
Importantly, our proposed approach can be used in general applications with weak supervision from multiple sources without any restrictive assumptions specific to crowdsourcing, and we show that our approach outperforms the aforementioned methods on a crowdsourcing benchmark task.

\section{End-to-End Weak Supervision}
\label{Methods}
\begin{wrapfigure}{rh}{0.6\textwidth}
\vspace{-0.8cm}
    \begin{minipage}{0.6\textwidth}
        \begin{algorithm}[H] 
        \caption{\label{alg:main} \weasel: The proposed Weakly Supervised End-to-end Learning algorithm for learning from multiple weak supervision sources.}
\begin{algorithmic}
    \STATE \textbf{input:}
    batch size $n$, networks $e$, $f$, inverse temperatures $\tau_1, \tau_2$, noise-aware loss function $L$, class balance $P(y)$.
    \FOR{sampled minibatch $\{ z^{(k)} = (\kthFeatures, \kthLfs)\}_{k=1}^n$}
    \STATE \textbf{for all} $k\in \{1, \ldots, n\}$ \textbf{do}
        \STATE $~~~~$
        \textcolor{gray}{\# Produce accuracy scores for all weak sources}
        \STATE $~~~~$
        $\theta\brackets{z^{(k)}}
        = \softmax\brackets{e(z^{(k)}) \tau_1}$
        \STATE $~~~~$
        \textcolor{gray}{\# Generate probabilistic labels}
        \STATE $~~~~$
        \textbf{define} $\kthScore$ \textbf{as}~ 
        $\kthScore = 
        \theta(z^{(k)})^T \lambdaOnehot^{(k)}$
        \STATE $~~~~$
        $y^{(k)}_e = P_\theta(y | \lfs^{(k)})
        = \softmax\brackets{ \kthScore  \tau_2} \odot P(y)$
        \STATE $~~~~$
        \textcolor{gray}{\# Downstream model forward pass}
        \STATE $~~~~$
        $ y^{(k)}_f 
        = f(\kthFeatures) $ 
    \STATE \textbf{end for}
    \STATE 
    $\mathcal{L}_f
    = \frac{1}{n} \sum_{k=1}^n
    L\brackets{y^{(k)}_f, \texttt{stop-grad}\brackets{ y^{(k)}_e}}$
    
    \STATE
    $\mathcal{L}_e
    = \frac{1}{n} \sum_{k=1}^n
    L\brackets{y^{(k)}_e, \texttt{stop-grad}\brackets{y^{(k)}_f}}$
    \STATE update $e$ to minimize $\mathcal{L}_e$, and $f$ to minimize $\mathcal{L}_f$
    \ENDFOR
    \STATE \textbf{return} downstream network $f(\cdot)$
            \end{algorithmic}
        \end{algorithm}
    \end{minipage}
\vspace{-1.cm}
\end{wrapfigure}

In this section we present our flexible base algorithm that we call \weasel, which can be extended to probabilistic sources and other network architectures (Section~\ref{Extension}). See Algorithm~\ref{alg:main} for its pseudocode. 

\subsection{Problem Setup} \label{ProblemSetup}
Let $(\features,y) \sim \mathcal{D}$ be the data generating distribution, where the unknown labels belong to one of $C$ classes: $y\in \Y = \{1, ..., C\}$.
As in~\cite{DP}, users provide an unlabeled training set $\mathbf{X} = \{ \mathbf{x}_i\}_{i=1}^N$, and $m$ labeling functions $\lfs = \lfs( \mathbf{x}) \in \{0, 1, ..., C \}^m$, where $0$ means that the LF abstained from labeling for any class.
We write $\lambdaOnehot= \brackets{\indicator{\lfs = 1}, \dots,  \indicator{\lfs = C}}\in \{0,1\} ^{m \times C}$ for the one-hot representation of the LF votes provided by the $m$ LFs for $C$ classes. 
Our goal is to train a downstream model $f: \mathcal{X} \rightarrow \Y$ on a \emph{noise-aware} loss $L(y_f, y_e)$ that operates on the model's predictions $y_f = f(\features)$ and \emph{probabilistic labels} $ y_e$ generated by an encoder model $e$ that has access to LF votes, $\lfs$, and features, $\features$. Note that prior work restricts the probabilistic labels to only being estimated from the LFs.
%
\subsection{Posterior Reparameterization}
Previous PGM based approaches assume that the joint distribution $p(\lfs, y)$ of the LFs and the latent true label can be modeled as a Markov Random Field (MRF) with pairwise dependencies between weak supervision sources \cite{DP, Snorkel, Multitask, triplets, TripletsMean}. These models are parameterized by a set of LF accuracy and intra-LF correlation parameters and in some cases by additional parameters to model LF and class label propensity.
Note however, that the aforementioned models ignore features $X$ when modeling the latent labels 
and therefore disregard that LFs may differ in their accuracy across samples and data slices.
 
We relax these assumptions, and instead view the latent label as an \emph{aggregation of the LF votes that is a function of the entire set of LF votes and features, on a sample-by-sample basis}.
That is, we model the probability of a particular sample $\features$ having the class label $c \in \Y$ as
\begin{align}
    P_\theta(y = c | \lfs) 
    &= \softmax \brackets{ \mathbf{s} }_c P(y = c),
    \label{eq:softlabels} \\
    \mathbf{s}
    &= \theta(\lfs,  \features)^T \lambdaOnehot \in \R^C.
\end{align}
where $\theta(\lfs,  \mathbf{x}) \in \R^m$ weighs the LF votes on a sample-by-sample basis and the softmax for class $c$ on $s$ is defined as
\begin{equation*}
    \softmax \brackets{ \mathbf{s} }_c
    =
    \frac
    {\exp\brackets{\theta(\lfs,  \features)^T\indicator{\lfs = c}}}
    {\sum_{j=1}^C\exp\brackets{\theta(\lfs,  \features)^T\indicator{\lfs = j}}}.
\end{equation*}
While we do not use the class balance $P(y)$ in our experiments for our own model, \weasel, it is frequently assumed to be known \cite{Multitask, triplets, TripletsMean}, and can be estimated from a small validation set, or  using LF outputs as described in \cite{Multitask}. 
Our formulation can be seen as a reparameterization of the posterior of the pairwise MRFs in \cite{Snorkel, Multitask, triplets}, where $\theta$ corresponds to the LF accuracies that are fixed across the dataset and are solely learned via LF agreement and disagreement signals, ignoring the informative features. 
We further motivate this formulation and expand upon this connection in the appendix \ref{sec:appendixReparam}.

\subsection{Neural Encoder}
Based on the setup introduced in the previous section and captured in Eq.~(\ref{eq:softlabels}), our goal is to estimate latent labels by means of learning sample-dependent accuracy scores $\theta(\lfs, \features)$, which we propose to parameterize by a neural encoder \encoder. This network takes as input the features $\features$ and the corresponding LF outputs $\lfs(\features)$ for a data point, and outputs unnormalized scores $e(\lfs, \features)\in \R^{m}$.
Specifically, we define
\begin{equation}
    \theta(\lfs, \features) = \tau_2 \cdot \softmax \brackets{e(\lfs, \features) \tau_1},
    \label{eq:AccuracyScores}
\end{equation}
where $\tau_2$ is a constant factor that scales the final softmax transformation in relation to the number of LFs $m$, and is equivalent to an inverse temperature for the output softmax in Eq. \ref{eq:softlabels}. It is motivated by the fact that most LFs are sparse in practice, and especially when the number of LFs is large this leads to small accuracy magnitudes without scaling (since, without scaling, the accuracies after the softmax sum up to one)\footnote{In our main experiments we set $\tau_2=\sqrt{m}$.}. $\tau_1$ is an inverse temperature hyperparameter that controls the smoothness of the predicted accuracy scores: The lower $\tau_1$ is, the less emphasis is given to a small number of LFs --  as $\tau_1 \rightarrow 0$, the model aggregates according to the equal weighted vote. 
The $\softmax$ transformation naturally encodes our understanding of wanting to aggregate the weak sources to generate the latent label. 

\subsection{Training the Encoder}
The key question now is how to train \encoder, i.e. how can we learn an accurate mapping of the sample-by-sample accuracies, given that we do not observe any labels? 

First, note that initializing \encoder  with random weights will lead to latent label estimates close to an equal weighted vote, which acts as a reasonable baseline for label models in data programming (and crowdsourcing), where in expectation votes of LFs are assumed to better than random guesses. Thus, $P_\theta(y |\lfs, \features)$ will provide a better than random initial guess for $y$.
\emph{We hypothesize that in most practical cases, features, latent label, and labeling function aggregations are intrinsically correlated due to the design decisions made by the users defining the features and LFs. Thus, we can jointly optimize $e$ and $f$  by maximizing their agreement with respect to the target downstream loss $L$ in an end-to-end manner}. 
See Algorithm~\ref{alg:main} for pseudocode of the resulting \weasel\  algorithm.
The natural classification loss is the cross-entropy, which we use in our experiments, but in order to encode our desire of maximizing the agreement of the two separate models that predict based on different views of the data, we adapt it\footnote{This holds for any asymmetric loss, while for symmetric losses this is not needed.} in the following form:
The loss is symmetrized in order to compute the gradient of both models using the other model's predictions as targets.
To that end, it is crucial to use the \texttt{stop-grad} operation on the targets (the second argument of $L$), i.e. to treat them as though they were ground truth labels.
This choice is supported by our synthetic experiment and ablations.
This operation has also been shown to be crucial in siamese, non-contrastive, self-supervised learning, both empirically \cite{BYOL, simSiam} and theoretically \cite{tian2021understanding}.
By minimizing simultaneously, both, $L(y_e, y_f)$ and $L(y_f, y_e)$ to jointly learn the network parameters for $e$ and the downstream model $f$ respectively, we learn the accuracies of the noisy sources $\lfs$ that best explain the patterns observed in the data, and vice versa the feature-based predictions that are best explained by aggregations of LF voting patterns.
\subsection{WeaSEL Design Choices}
\vspace{-2mm}
Note that it is necessary to encode the inductive bias that the unobserved ground truth label $y$ is a (normalized) linear combination of LF votes -- weighted by sample- and feature-dependent accuracy scores. Otherwise, if the encoder network directly predicts $P_\theta(y |\lfs, \features)$ instead of the accuracies $\theta(\lfs, \features)$, the pair of networks $e, f$ have no incentive to output the desired latent label, without observed labels.  
We do acknowledge that this two-player cooperation game with strong inductive biases could still allow for degenerate solutions. However, we empirically show that our simple \weasel\  model that goes beyond multiple earlier WS assumptions is 1) competitive and frequently outperforms state-of-the-art PGM-based and crowdsourcing models (see Tables \ref{tab:mainResults} and \ref{tab:labelMe}); and 2) is robust against massive LF correlations and able to recover the performance of a fully supervised model on a synthetic example, while all other models break in this setting (see section \ref{sec:mainRobustness} and appendix \ref{sec:robustnessAppendix}).  
%


\begin{table*} 
    \centering
    \caption{The final test F1 performance of various multi-source weak supervision methods over seven runs, using different random seeds, are averaged out $\pm$ standard deviation. 
    The top 2 performance scores are highlighted as \first{First}, \second{Second}. 
    Triplet-median~\cite{TripletsMean} is not listed  as it only converged for IMDB with 12 LFs (F1 = $73.0 \pm 0.22$), and Spouse (F1 = $48.7 \pm 1.0$).
    The downstream model is the same for all methods. For
    Sup. (Val. set), and Majority vote it is trained on the hard labels induced by the labeled validation set and the majority vote of the LFs, respectively. For the rest it is trained on the probabilistic labels estimated by the respective state-of-the-art latent label model.
    For reference, we also report the \emph{Ground truth} performance of the same downstream model trained on the true training labels (which are unused by all other models, and not available for Spouse).
    } 
    \vspace{5pt}
    \scalebox{0.8}{
    \begin{tabular}{c|ccccccccc|}
        \toprule
        \multicolumn{1}{c|}{\textbf{Model}} &
        \multicolumn{1}{c}{\textbf{Spouse} \small{(9 LFs)}} &
        \multicolumn{1}{c}{\textbf{ProfTeacher} \small{(99 LFs)}} &
        \multicolumn{1}{c}{\textbf{IMDB} \small{(136 LFs)}} &
        \multicolumn{1}{c}{\textbf{IMDB} \small{(12 LFs)}}  &
        \multicolumn{1}{c}{\textbf{Amazon} \small{(175 LFs)}} &
        \\
        \midrule
        \midrule
        Ground truth &
            -- &
            $90.65 \pm 0.29$ &
            $86.72 \pm 0.40$ &
            $86.72 \pm 0.40$ &
            $92.93 \pm 0.68$ \\ 
         \cline{1-6}
        Sup. (Val. set) &
            $20.4 \pm 0.2$ &
            $73.34 \pm 0.00$ &
            $68.76 \pm 0.00$ &
            $68.76 \pm 0.00$ &
            $84.18 \pm 0.00$ \\ 
        Snorkel &        
            $48.79 \pm 2.69$  &
            $85.12 \pm 0.54$  &
            \first{82.22 $\pm$ 0.18} &
            \second{74.45 $\pm$ 0.58} &
            $80.54 \pm 0.41$ \\
        Triplet &
            $45.88 \pm 3.64$ &
            $74.43 \pm 10.59$ & 
            $75.36 \pm 1.92$ &
            $73.15 \pm 0.95$ &
            $75.44 \pm 3.21$ \\
        Triplet-Mean &    
            \second{49.94 $\pm$ 1.47} &
            $82.58 \pm 0.32$ &
            $79.03 \pm 0.26$ &
            $73.18 \pm 0.23$ &
            $79.44 \pm 0.68$ \\
        Majority vote &
            $40.67 \pm 2.01$ &
            \second{85.44 $\pm$ 0.37} &
            $80.86 \pm 0.28$ &
            $74.13 \pm 0.31$ &
            \second{84.20 $\pm$ 0.52} \\
        \cline{1-6}
        \weasel  & 
            \first{51.98 $\pm$ 1.60} &
            \first{86.98 $\pm$ 0.45} &
            \second{82.10 $\pm$ 0.45} &
            \first{77.22 $\pm$ 1.02} &
            \first{86.60 $\pm$ 0.71}\\ 
        \bottomrule
    \end{tabular}
    \label{tab:mainResults}
}
\end{table*}

\section{Experiments}
\label{Experiments}
\textbf{Datasets} $\;$
As in related work on label models for weak supervision~\cite{DP, Multitask, triplets, TripletsMean}, we focus for simplicity on the binary classification case with unobserved ground truth labels $y \in \{-1, 1\}$.
See Table~\ref{tab:datasets} for details about dataset sizes and the number of LFs used.
We also run an experiment on a multi-class, crowdsourcing dataset (see subsection \ref{sec:crowdsourcing}).
We evaluate the proposed end-to-end system for learning a downstream model from multiple weak supervision sources on previously used benchmark datasets in weak supervision work~\cite{Snorkel, IWS, TripletsMean}.   Specifically, we evaluate test set performance on the following classification datasets:

\begin{itemize}
    \item 
    \textit{The IMDB movie review} dataset \cite{IMDB} contains  movie reviews to be classified into positive and negative sentiment. We run two separate experiments, where in one we use the same 12 labeling functions as in \cite{TripletsMean}, and for the other we choose 136 text-pattern based LFs. More details on the LFs can be found in the appendix \ref{sec:implDetails}.
    \item
    A subset of the \textit{Amazon review} dataset \cite{Amazon}, where the task is to classify product reviews into positive and negative sentiment.
    \item
    We use the \textit{BiasBios biographies} dataset \cite{biasBios} to distinguish between binary categories of frequently occurring occupations and use the same subset of professor vs teacher classification as in~\cite{IWS}.
    \item
    Finally, we use the highly unbalanced \textit{Spouse} dataset (90\% negative class labels), where the task is to identify mentions of spouse relationships amongst a set of news articles from the Signal Media Dataset \cite{Spouses}.
\end{itemize}
For the Spouse dataset, the same data split and LFs as in \cite{triplets} are used, while for the rest we take a small subset of the 
test set as validation set. This is common practice in the related work \cite{Snorkel, Multitask, triplets, IWS} for tuning hyperparameters, and allows for a fair comparison of models.

\subsection{Benchmarking Weak Supervision Label Models}
To evaluate the proposed system, we benchmark it against state-of-the-art systems that aggregate multiple weak supervision sources for classification problems, without any labeled training data.
We compare our proposed approach with the following systems: 1) \emph{Snorkel}, a popular system proposed in \cite{Snorkel, Multitask}; 2) \emph{Triplet}, exploits a closed-form solution for binary classification under certain assumptions \cite{triplets}; and 3) \emph{Triplet-mean} and \emph{Triplet-median}~\cite{TripletsMean}, which are follow-up methods based on \emph{Triplet} with the aim of making the method more robust.

We report the held-out test set performance of \weasel's downstream model $f$. Note that in many settings it is often not possible to apply the encoder model to make predictions at test time, since the LFs usually do not cover all data points (e.g. in Spouse only 25.8\% of training samples get at least one LF vote), and can be difficult to apply to new samples (e.g. when the LFs are crowdsourced annotations). In contrast, the downstream model is expected to generalize to arbitrary unseen data points.

We observe strong results for our model, with 4 out of 5 top scores, and a lift of 6.1 F1 points over the next best label model-based method in the Amazon dataset. Our results are summarized in Table~\ref{tab:mainResults}.
Since our model is based on a neural network, we hypothesize that the large relative lift in performance on the Amazon review dataset is due to it being the largest dataset size on which we evaluate on -- we expect this lift to hold or become larger as the training set size increases. 
To obtain the comparisons shown in Table~\ref{tab:mainResults}, we run Snorkel over six different label model hyperparameter configurations, and train the downstream model on the labels estimated by the label model with the best AUC score on the validation set.  We do not report Triplet-median in the main table, since it only converged for the two tasks with very small numbers of labeling functions. 
Interestingly, we observed that training the downstream model on the hard labels induced by majority vote leads to a competitive performance, better than triplet methods in four out of five datasets. This baseline is not reported in previous papers (only the raw majority vote is usually reported, without training a classifier). Our own model, \weasel, on the other hand consistently improves over the majority vote baseline (which in Table \ref{tab:tableResultsAppendix}, in the appendix, can be seen to lead to similar performance as an untrained encoder network, $e$, that is left at its random initialization).
\subsection{Crowdsourcing dataset} \label{sec:crowdsourcing}
          
\begin{wraptable}{r}{6cm}
\vspace{-4mm}
\caption{Test accuracy scores on the crowd-sourced, multi-class LabelMe image classification dataset.
}
\vspace{-3mm}
\begin{center}
\begin{tabular}{llllllllll}
\toprule
    Model & Accuracy \\ 
    \midrule
        Majority vote        & $79.23 \pm 0.5$\\  
        MBEM \cite{khetan2017learning} & $76.84 \pm 0.4$ \\
        DoctorNet \cite{doctorNet} & $81.31 \pm 0.4$\\ 
        CrowdLayer \cite{rodrigues2018deep} & $82.83 \pm 0.4$\\ 
        AggNet \cite{aggNet} & $84.35 \pm 0.4$ \\
        MaxMIG \cite{MaxMIG} & \second{85.45 $\pm$ 1.0}\\ 
        Snorkel+CE & $82.89 \pm 0.7$\\ 
        \weasel+CE & $82.46 \pm 0.8$\\             
        Snorkel+MIG & $85.15 \pm 0.8$\\ 
        \weasel+MIG & \first{86.36 $\pm$ 0.3}\\     
    \bottomrule
    \label{tab:labelMe}
\end{tabular}
\end{center}
\vspace{-5mm}
\end{wraptable}
Data programming and crowdsourcing methods have been rarely compared against each other, even though the problem setup is quite similar.
Indeed, end-to-end systems specifically for crowdsourcing have been proposed \cite{raykar10a, khetan2017learning, rodrigues2018deep, MaxMIG}. These methods follow crowdsourcing-specific assumptions and modeling choices (e.g. independent crowdworkers, a confusion matrix model for each worker, and in general build upon \cite{SkeneModel}). 
Still, since crowdworkers can be seen as a specific type of labeling functions, the performance of general WS methods on crowdsourcing datasets is of interest, but has so far not been studied. We therefore choose to also evaluate our method on the multi-class LabelMe image classification dataset that was previously used in the core related crowdsourcing literature \cite{rodrigues2018deep, MaxMIG}.
The results are reported in Table \ref{tab:labelMe}, and more details on this experiment can be found in Appendix \ref{sec:crowdsourcingAppendix}.
Note that the evaluation procedure in \cite{MaxMIG} reports the best test set performance for all models, while we follow the more standard practice of reporting results obtained by tuning based on a small validation set -- as in our main experiments.
We find that our model, \weasel, is able to outperform Snorkel as well as multiple state-of-the-art methods that were specifically designed for crowdsourcing (including several end-to-end approaches).
Interestingly, this is achieved by using the mutual information gain loss (MIG) function introduced in \cite{MaxMIG}, which significantly boosts performance of both Snorkel (the end-model, $f$, trained on the MIG loss with respect to soft labels generated by the first Snorkel label model step) and \weasel\ that use the cross-entropy (CE) loss.
This suggests that the MIG loss is a great choice for the special case of crowdsourcing, due to its strong assumptions common to crowdsourcing which are much less likely to hold for general LFs. This is reflected in our ablations too, where using the MIG loss leads to a consistently worse performance on our main multi-source weak supervision datasets.
\subsection{Robustness to Adversarial LFs and LF correlations}
\label{sec:mainRobustness}
\begin{figure}
    \begin{subfigure}{.5\textwidth}
      \centering
      \includegraphics[width=.95\linewidth]{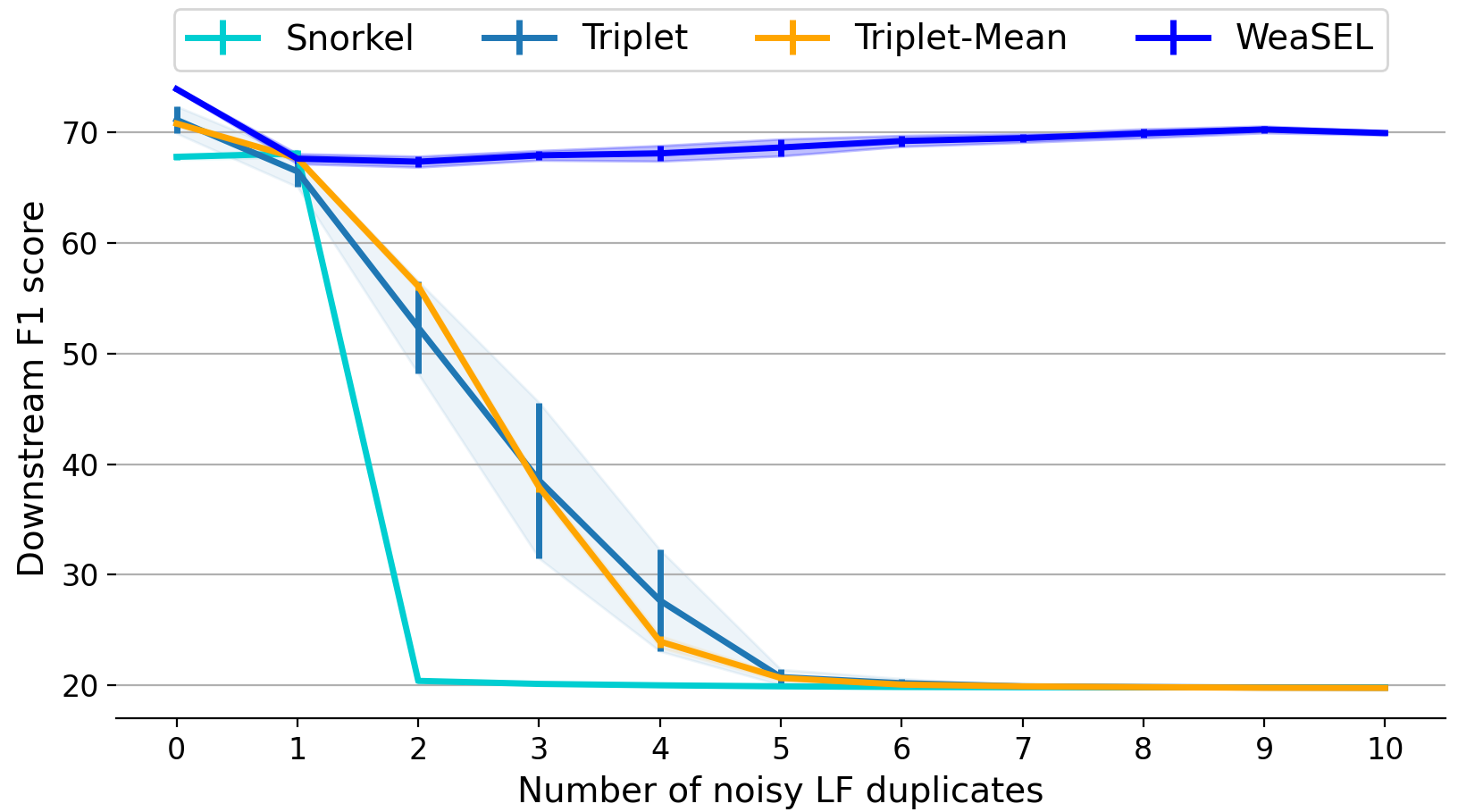}
      \caption{Test F1 score on robustness experiment \\ as a function of the number of adversarial LFs.}
      \label{fig:noisyLF}
    \end{subfigure}%
    \begin{subfigure}{.5\textwidth}
      \centering
      \includegraphics[width=.95\linewidth]{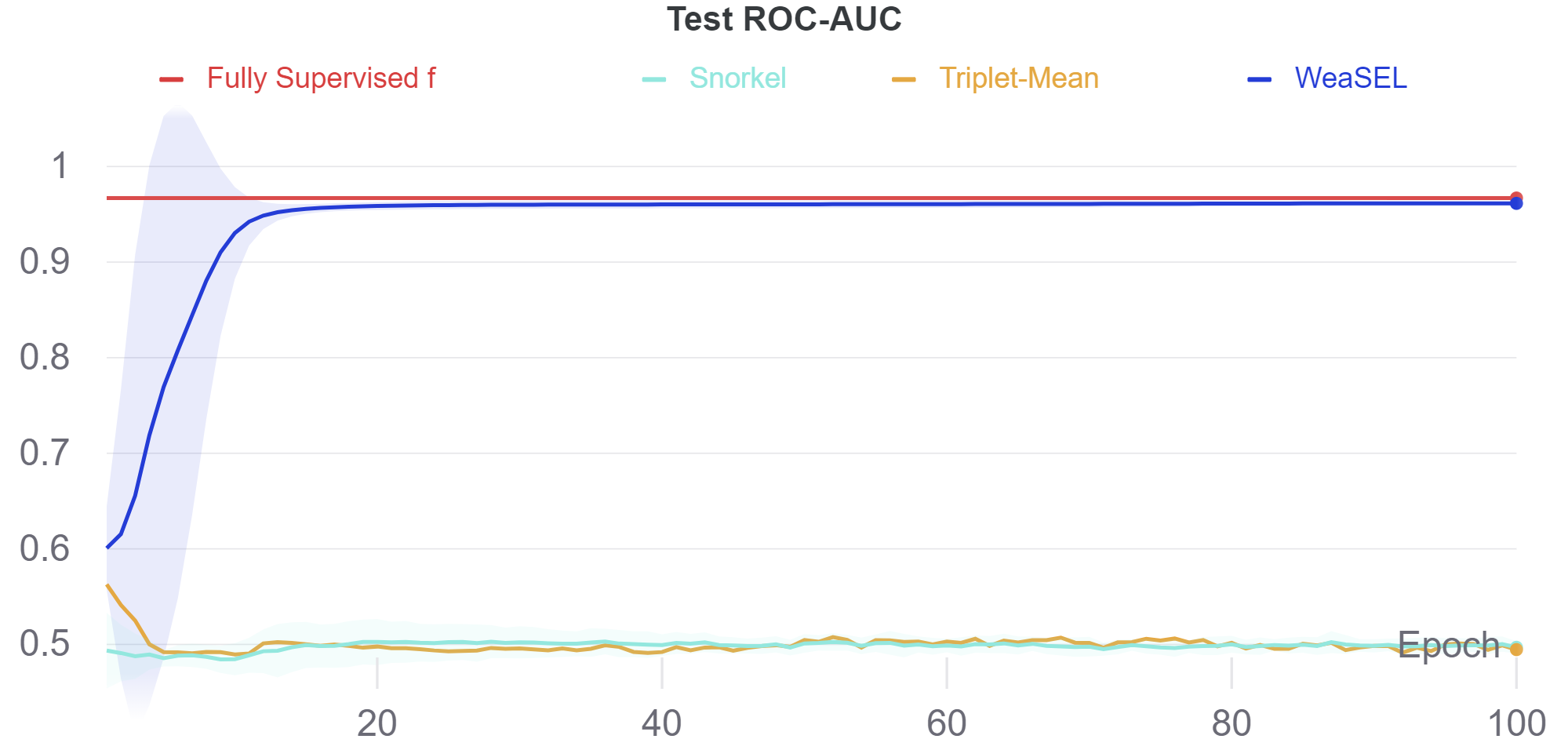}
      \caption{Test AUC by epoch in an experiment where one LF corresponds to the true class label and others are random.}
      \label{fig:SynthRobustnessTestCurve}
    \end{subfigure}
    \caption{
    \weasel\ is significantly more robust against correlated adversarial (left) or random (right) LFs than prior work whose assumptions make them equivalent to a Naive Bayes model.
    For subfigure (a), we duplicate a fake adversarial LF up to 10 times, and observe that our end-to-end system is robust against the adversarial LF, while other systems quickly degrade in performance (over ten random seeds). 
    In (b), we let one LF be the true labels $y^*$ and then duplicate a LF that votes according to a coin flip 2, 5, ..., 2000 times. We plot the test AUC performance curve as a function of the epochs, averaged out over the different number of duplications (and five random seeds). \weasel\ consistently recovers the test performance of the supervised end-model $f$ trained directly on the true labels $y^*$, whose end performance (AUC $=0.967$) is shown in red. 
    }
  \label{fig:mainRobustness}
\end{figure}

Users will sometimes generate sources they mistakenly think are accurate. This also encompasses the 'Spammer' crowdworker-type studied in the crowdsourcing literature.
Therefore, it is desirable to build models that are robust against such sources.
We argue that our system that is trained by maximizing the agreement between an aggregation of the sources and the downstream model's predictions should be able to distinguish the adversarial sources.
In Fig.~\ref{fig:noisyLF} we show that our system does not degrade in its initial performance, even after duplicating an adversarial LF ten times.
Prior latent label models, on the other hand, rapidly degrade, given that they often assume the weak label sources to be conditionally independent given the latent label, equivalent to a Naive Bayes generative model.
Note that the popular open-source implementation of \cite{Snorkel, Multitask}
does not support user-provided LF dependencies modeling, while \cite{triplets, TripletsMean} did not converge in our experiments when modeling dependencies, and as such we were not able to test their performance when the correlation dependencies between the duplicates are provided (which in practice, of course, are not known).

We also run a synthetic experiment inspired by \cite{MaxMIG}, where one LF is set to the true labels of the ProfTeacher dataset, i.e. $\lambda_1 = y^*$, while the other LF simply votes according to a coin flip, i.e. $\lambda_2 \sim P(y)$, and we then duplicate this latter LF, i.e. $\lambda_3 = \dots = \lambda_m=\lambda_2$. Under this setting, our \weasel\ model is able to consistently \emph{recover the fully supervised performance} of the same downstream model directly trained on the true labels $y^*$, \emph{even when we duplicate the random LF up to $2000$ times} ($m=2001$). 
Snorkel and triplet methods, on the other hand, were unable to recover the true label (AUC $\approx0.5$).
Importantly, we find that the design choices for \weasel\ are to a large extent key in order to recover the true labels in a stable manner as in Fig.~\ref{fig:SynthRobustnessTestCurve}. Various other choices either collapse similarly to the baselines, are not able to fully recover the supervised performance, or lead to unstable test performance curves, see Fig. \ref{fig:syntheticRobustnessAppendix} in the appendix. 
More details on the experimental design and an extensive discussion, ablation, and figures based on the synthetic experiment can be found in the appendix \ref{sec:robustnessAppendix}.

\subsection{Implementation Details}
Here we provide a high-level overview over the used encoder architecture, the LF sets, and the features. More details, especially hyperparameter and architecture details, are provided in Appendix \ref{sec:implDetails}. All downstream models are trained with the (binary) cross-entropy loss, and our model with the symmetric version of it that uses \texttt{stop-grad} on the targets.

\textbf{Encoder network} $\;$
The encoder network $e$ does not need to follow a specific neural network architecture and we therefore use a simple multi-layer perceptron (MLP) in our benchmark experiments. 

\textbf{Features for the encoder} $\;$
A big advantage of our model is that it is able to take into account the features $\features$ for generating the sample-by-sample source accuracies. For all datasets, we concatenate the LF outputs with the same features that are used by the downstream model as input of our encoder model (for Spouse we use smaller embeddings than the ones used by the downstream LSTM). 

\textbf{Weak supervision sources} $\;$
For the Spouse dataset, and the IMDB variant with 12 LFs, we use the same LFs as in \cite{triplets, TripletsMean} respectively. The remaining three LF sets were selected by us prior to running experiments.
These LFs are all pattern- and regex-based heuristics, while the Spouse experiments also contain LFs that are distant supervision sources based on DBPedia. 

\begin{table}
\centering
\caption{Dataset details, where training, validation and test set sizes are $N_{train}$, $N_{val}$, $N_{test}$ respectively, and $f$ denotes
the downstream model type. We also report the total coverage Cov. of all LFs, which refers to the percentage of training samples which are labeled by at least one LF (the rest is not used). For IMDB we used two different sets of labeling functions of sizes 12 and 136. }
\begin{tabular}{@{} *7l @{}}
\toprule
    Dataset  & \#LFs & $N_{train}$ & Cov. \small{(in \%)}  & $N_{val}$  & $N_{test}$ & $f$\\ 
    \midrule
    Spouse  & $9$     & $22,254$  & $25.8$ & $2811$ & $2701$    & LSTM \\
    BiasBios & $99$    & $12,294$ & $81.8$ & $250$  & $12,044$  & MLP \\
    IMDB     & $12$    & $25k$    & $88.0$ & $250$  & $24,750$     & MLP \\
    IMDB     & $136$    & $25k$   & $83.1$ & $250$  & $24,750$     & MLP \\
    Amazon   & $175$    & $160k$  & $65.5$    & $500$  & $39,500$     & MLP \\
\bottomrule
\end{tabular}
\label{tab:datasets}
\end{table}

\section{Ablations}
In this section we demonstrate the strength of the \weasel\ model design decisions. We perform extensive ablations on all four main datasets but Spouse for twenty configurations of \weasel\ with different encoder architectures, hyperparameters, and loss functions.
The tabular results and a more detailed discussion than in the following can be found in Appendix \ref{sec:ablationsAppendix}.

We observe that ignoring the features when modeling the sample-dependent accuracies, i.e. $\theta(\lfs, \features) = \theta(\lfs)$, usually underperforms by up to 1.2 F1 points.
A more drastic drop in performance, up to 4.9 points, occurs when the encoder network is linear, i.e. without hidden layers, as in \cite{MaxMIG}.
It also proves helpful to scale the $\softmax$ in Eq. \ref{eq:AccuracyScores} by $\sqrt{m}$ via the inverse temperature parameter $\tau_2$. Further, while the MIG loss proved important for \weasel\ to achieve state-of-the-art performance on the crowdsourcing dataset (with a similar lift in performance observable for Snorkel using MIG for downstream model training), this does not hold for the main datasets. This indicates that the MIG loss is a good choice for crowdsourcing, but not for more general WS settings.  

Our ablations also show that it is important to restrict the accuracies to a positive interval (e.g. (0, 1), with the sigmoid function being a good alternative to the softmax we use). On the one hand, this encodes the inductive bias that LFs are not adversarial, i.e. can not have negative accuracies, (using tanh to output accuracy scores does not perform well), and on the other hand does not give the encoder network too much freedom in the scale of the scores (using ReLU underperforms significantly as well). 

Additionally, we find that our choice of using the symmetric cross-entropy loss with \texttt{stop-grad} applied to the targets is  crucial for the obtained strong performance of \weasel. Removing the \texttt{stop-grad} operation, or using the standard cross-entropy (without \texttt{stop-grad} on the target) leads to significantly worse scores and a very brittle model.
Losses that already are symmetric (e.g. L1 or Squared Hellinger loss) neither need to be symmetrized nor use \texttt{stop-grad}. While the L1 loss consistently underperforms, we find that the Squared Hellinger loss can lead to better performance on two of the four datasets.

However, only the symmetric cross-entropy loss with \texttt{stop-grad} on the targets is shown to be robust and able to recover the true labels in our synthetic experiment in Section \ref{sec:mainRobustness}. Thus, to complement the above ablation on real datasets, we additionally run extensive ablations on this synthetic setup in Appendix \ref{sec:robustnessAppendix}. This synthetic ablation gives interesting insights, and strongly supports the proposed design of \weasel. Indeed, many choices for \weasel\ that perform well enough on the real datasets, such as no features for the encoder, $\tau_2 = 1$, sigmoid parameterized accuracies, and all other losses that we evaluated, lead to significantly worse performance and less robust learning on the synthetic adversarial setups.

\section{Practical Aspects and Limitations}
\label{sec:practicalAspects}
\paragraph{On why it works \& degenerate solutions}
Overall, \weasel\ avoids trivial overfitting and degenerate solutions by hard-coding the encoder generated labels as a (normalized) linear combination of the $m$ LF outputs, weighted by $m$ sample-dependent accuracy scores. 
This design choice also ensures that the randomly initialized $e$ will lead the downstream model $f$ that is trained on soft labels generated by the random encoder, to obtain performance similar to when $f$ is trained on majority vote labels. In fact, the random-encoder-\weasel\ variant itself often outperforms other baselines, and triplet methods in particular (see appendix \ref{sec:resultsAppendix}).

Empirically, we only observed degenerate solutions when training for too many epochs. Early-stopping on a small validation set ensures that a strong final solution is returned, and should be done whenever such a set exists or is easy to create. When no validation set is available, we find that choosing the temperature hyperparameter in Eq. \ref{eq:AccuracyScores} such that $\tau_1\le 1/3$ avoids collapsed solutions on all our datasets. This can be explained by the fact that a lower inverse temperature forces the encoder-predicted label to always depend on multiple LF votes when available, rather than a single one (which happens when the $\softmax$ in Eq.~\ref{eq:AccuracyScores} becomes a $\texttt{max}$ as $\tau_1 \rightarrow \infty$). This makes it harder for the encoder to overfit to individual LFs. Our ablations indicate that this temperature parameter setting comes at a small cost in terms of loss in downstream performance, compared to when using a validation set for early stopping. Thus, when no validation set is available, we advise to lower $\tau_1$.

\paragraph{Complex downstream models}
We have shown that \weasel\ achieves competitive or state-of-the-art performance on all datasets we tried it on, for a given set of LFs. In practice, however, this LF set needs to first be defined by users. This can be done via an iterative process, where the feedback is sourced from the quality of the probabilistic labels generated by the label model. A limitation of our model, is that each such iteration would require training the downstream model, $f$. When $f$ is slow to train, this may slow down the LF development cycle and lead to unnecessary energy consumption. A practical solution to this can be to a) do the iteration cycle with a less complex downstream model; or b) use the fast to train PGM-based label models to choose a good LF set, and then move to \weasel\ in order to achieve better downstream performance.

\section{Extensions}
\label{Extension}
\paragraph{Probabilistic labeling functions}
Our learning method can easily support labeling functions that output continuous scores instead of discrete labels as in \cite{chatterjee2019data}. In particular, this includes probabilistic sources that output a distribution over the potential class labels.
This can be encoded in our model by changing the one-hot representation of our base model to a continuous representation $\lambdaOnehot \in [0, 1]^{m \times C}$.

\paragraph{Modeling more structure}
While we use a simple multi-layer perceptron (MLP) as our encoder $e$ in our benchmark experiments, our formulation is flexible to support arbitrarily complex networks. In particular, we can naturally model dependencies amongst weak sources via edges in a Graph Neural Network (GNN), where each LF is represented by a node that is given the LF outputs as features.
Furthermore, while  we only explicitly reparameterized  the accuracy parameters of the sources in our base model, it is straightforward to augment $\lambdaOnehot$ with additional sufficient statistics, e.g. the fixing or priority dependencies from \cite{DP, MisspecificationInDP} that encode that one source fixes (i.e. should be given priority over) the other whenever both vote.

\section{Conclusion}
We proposed \weasel, a new approach for end-to-end learning of  neural network models for classification from, exclusively, multiple sources of weak supervision that streamlines prior latent variable models. 
We evaluated the proposed approach on benchmark datasets and observe that the downstream models outperform state-of-the-art data programming approaches in 4 out of 5 cases while remaining highly competitive on the remaining task, and outperforming several state-of-the-art crowdsourcing methods on a crowdsourcing task. 
We also demonstrated that our integrated approach can be more robust to dependencies between the labeling functions as well as to adversarial labeling scenarios. 
The proposed method works with discrete and probabilistic labeling functions and can utilize various neural network designs for probabilistic label generation. 
This end-to-end approach can simplify the process of developing effective machine learning models using weak supervision as the primary source of training signal, 
and help adoption of this form of learning in a wide range of practical applications.


\newpage
\section*{Acknowledgements}
This work was made possible thanks to Carnegie Mellon University's Robotics Institute Summer Scholars program and was partially supported by a Space Technology Research Institutes grant from NASA’s Space Technology Research Grants Program, and by Defense Advanced Research Projects Agency's award FA8750-17-2-0130.

\bibliography{references}
\bibliographystyle{plainnat} 

\clearpage
\appendix
\section*{Appendix}
\section{Posterior Reparameterization}
\label{sec:appendixReparam}
In this section we motivate the design choices and inductive biases that we encode into our neural encoder network $e$, which is the network that is used to model the relative accuracies of the weak supervision sources $\lfs$. 
%
Recall that we model the probability of a particular sample $\features \in \mathcal{X}$ having the class label $y \in \Y = \{1, \dots, C\}$ as
\begin{align}
    P_\theta(y | \lfs) 
    &= \softmax \brackets{ \mathbf{s} }_y P(y),
    \label{eq:softlabelsAppendix} \\
    \mathbf{s}
    &= \theta(\lfs,  \features)^T \lambdaOnehot \in \R^C.
\end{align}
where $\theta(\lfs,  \mathbf{x}) \in \R^m$ weighs the LF votes on a sample-by-sample basis and the softmax for class $y$ on $s$ is defined as
\begin{equation*}
    \softmax \brackets{ \mathbf{s} }_y
    =
    \frac
    {\exp\brackets{\theta(\lfs,  \features)^T\indicator{\lfs = y}}}
    {\sum_{y' \in \Y}\exp\brackets{\theta(\lfs,  \features)^T\indicator{\lfs = y'}}}.
\end{equation*}
\textbf{Connection to prior PGM models} $\;$
We now motivate this choice by deriving a less expressive variant of it from the standard Markov Random Field (MRF) used in the related work.
If we view the attention scores $\theta(\lfs,  \features) \in \R^m$, that assign sample-dependent accuracies to each labeling function, as sample-independent parameters $\theta_1$ and, by that, drop the features from the equation -- as is done in the related work \cite{DP, Multitask, triplets, TripletsMean} -- we can rewrite Eq.~\ref{eq:softlabelsAppendix} as 
\begin{align*}
    &     
    \frac
    {\exp\brackets{\theta_1^T\indicator{\lfs = y}}}
    {\sum_{y' \in \Y}\exp\brackets{\theta_1^T\indicator{\lfs = y'}}} P(y) \\
\intertext{Let $\phi_1(\lfs, y) = \indicator{\lfs = y}$, and, for clarity of writing, we drop the class balance, then this becomes}
    &= 
    \frac
    {\exp\brackets{\theta_1^T\phi_1(\lfs, y)}}
    {\sum_{y' \in \Y}\exp\brackets{\theta_1^T\phi_1(\lfs, y')}} \\
    &= 
    \frac
    {Z_\theta^{-1}\exp\brackets{\theta_1^T\phi_1(\lfs, y) + \theta_2^T\phi_2(\lfs)}}
    {\sum_{y' \in \Y}Z_\theta^{-1}\exp\brackets{\theta_1^T\phi_1(\lfs, y') + \theta_2^T\phi_2(\lfs)}} \\
    &=
    \frac
    {P_\theta \brackets{\lfs, y}}
    {\sum_{y' \in \Y} P_\theta \brackets{\lfs, y'}} \\
    &= 
    \frac
    {P_\theta \brackets{\lfs, y}}
    {P_\theta \brackets{\lfs}} \\
    &= 
    P_\theta \brackets{y | \lfs},
\end{align*}
where in the second step we multiplied the denominator and numerator with the same quantity $\frac{1}{Z_\theta}\exp\brackets{\theta_2^T\phi_2(\lfs)}$, and $\theta$ now parameterizes the joint distribution of the latent label and weak sources as 
$$
P_\theta(\lfs, y) = \frac{1}{Z_\theta} \exp\brackets{\theta_1^T\phi_1(\lfs, y) + \theta_2^T\phi_2(\lfs)} = \frac{1}{Z_\theta} \exp\brackets{\theta^T\phi(\lfs, y)}.
$$ 
We can recognize $P_\theta$ as a distribution from the exponential familiy, and more specifically as a pairwise MRF, or factor graph, with canonical parameters $\theta = \brackets{\theta_1, \theta_2}$ and corresponding sufficient statistics, or factors, $\phi(\lfs, y) = \brackets{\phi_1(\lfs, y), \phi_2(\lfs)}$, as well as the log partition function $Z_\theta$.
The accuracy factors and parameters $\phi_1, \theta_1$ are the core component of this model and sometimes take the form $\phi_1(\lfs y) = \lfs y$ in binary models as in \cite{DP, triplets, TripletsMean}.
The label-independent factors $\phi_2(\lfs)$ have, as can be seen from the derivation above, no direct influence on the latent label posterior, but are often used to model labeling propensities $\indicator{\lfs \neq 0}$ and correlation dependencies $\indicator{\lambda_i = \lambda_j}$, which can be important for PGM parameter learning, but are susceptible to misspecifications \cite{structureLearning2, TripletsMean, MisspecificationInDP}.
\emph{Our own parameterization therefore is a more expressive variant of these latent-variable PGM models, where we are able to assign LF accuracies on a sample-by-sample basis. Furthermore, our neural encoder network outputs them as a function of the LF outputs \textbf{and} features, and is expected to learn the easy to misspecify dependencies and label-independent statistics implicitly. Indeed, our empirical findings and subsection 4.3 support this.}

\begin{table*}[h!] 
    \centering
    \caption{The final test F1 performance of various multi-source weak supervision methods over seven runs, using different random seeds, are averaged out $\pm$ standard deviation. 
    The top 2 performance scores are highlighted as \first{First}, \second{Second}.
    Triplet-median~\cite{TripletsMean} is not listed  as it only converged for IMDB with 12 LFs (F1 = $73.0 \pm 0.22$), and Spouse (F1 = $48.7 \pm 1.0$). Sup. (Val. set) is the performance of the downstream model trained in a supervised manner on the labeled validation set. The rest are state-of-the-art latent label models.
    For reference, we also report the \emph{Ground truth} performance of a fully supervised model trained on true training labels (which are unused by all other models, and not available for Spouse).
    We also report the performance of \weasel-random, where only the downstream model of \weasel\ is trained (and the encoder network is left at its randomly initialized state).
    All models are run twice, where only the learning rate differs (either $10^{-4}$ or $4\cdot10^{-5}$), and the model with best ROC-AUC on the validation set is reported.
    The probabilistic labels from Snorkel used for downstream model training are chosen over six different configurations of the learning rate and number of epochs (again with respect to validation set ROC-AUC).
    } 
    \vspace{5pt}
    \scalebox{0.8}{
    \begin{tabular}{c|ccccccccc|}
        \toprule
        \multicolumn{1}{c|}{\textbf{Model}} &
        \multicolumn{1}{c}{\textbf{Spouse} \small{(9 LFs)}} &
        \multicolumn{1}{c}{\textbf{ProfTeacher} \small{(99 LFs)}} &
        \multicolumn{1}{c}{\textbf{IMDB} \small{(136 LFs)}} &
        \multicolumn{1}{c}{\textbf{IMDB} \small{(12 LFs)}}  &
        \multicolumn{1}{c}{\textbf{Amazon} \small{(175 LFs)}} &
        \\
        \midrule
        \midrule
        Ground truth &
            -- &
            $90.65 \pm 0.29$ &
            $86.72 \pm 0.40$ &
            $86.72 \pm 0.40$ &
            $92.93 \pm 0.68$ \\ 
         \cline{1-6}
        Sup. (Val. set) &
            $20.4 \pm 0.2$ &
            $73.34 \pm 0.00$ &
            $68.76 \pm 0.00$ &
            $68.76 \pm 0.00$ &
            $84.18 \pm 0.00$ \\ 
        Snorkel &        
            $48.79 \pm 2.69$  &
            $85.12 \pm 0.54$  &
            \first{82.22 $\pm$ 0.18} &
            \second{74.45 $\pm$ 0.58} &
            $80.54 \pm 0.41$ \\
        Triplet &
            $45.88 \pm 3.64$ &
            $74.43 \pm 10.59$ & 
            $75.36 \pm 1.92$ &
            $73.15 \pm 0.95$ &
            $75.44 \pm 3.21$ \\
        Triplet-Mean &    
            \second{49.94 $\pm$ 1.47} &
            $82.58 \pm 0.32$ &
            $79.03 \pm 0.26$ &
            $73.18 \pm 0.23$ &
            $79.44 \pm 0.68$ \\
       \cline{1-6}
         \weasel-random &
            $46.43 \pm 3.29$ &
            $83.47 \pm 0.64$ &
            $79.80 \pm 0.48$ &
            $74.22 \pm 0.45$ &
            $82.22 \pm 0.57$ \\
        Majority vote &
            $40.67 \pm 2.01$ &
            \second{85.44 $\pm$ 0.37} &
            $80.86 \pm 0.28$ &
            $74.13 \pm 0.31$ &
            \second{84.20 $\pm$ 0.52} \\
        \cline{1-6}
        \weasel  & 
            \first{51.98 $\pm$ 1.60} &
            \first{86.98 $\pm$ 0.45} &
            \second{82.10 $\pm$ 0.45} &
            \first{77.22 $\pm$ 1.02} &
            \first{86.60 $\pm$ 0.71}\\ 
        \bottomrule
    \end{tabular}
    \label{tab:tableResultsAppendix}
}
\end{table*}

\section{Extended Results} \label{sec:resultsAppendix}
We provide more detailed results in Table \ref{tab:tableResultsAppendix}.
Here, we include \weasel-random, which corresponds to \weasel\ with a randomly initialized encoder network that is not trained/updated. 
As expected, this setting produces performance often similar compared to training an end model on the hard majority vote labels. This is due to the strong inductive bias in our encoder model that constrains the encoder labels to be a normalized linear combination of the LF votes, weighted by positive accuracy scores.
In fact, \weasel-random itself is often able to outperform the PGM-based baselines, in particular the triplet methods.
Our results show that \weasel\ consistently improves significantly upon these baselines via training the encoder network to maximize its agreement with the downstream model.

\section{Extended Implementation Details}
\label{sec:implDetails}
\paragraph{Weak supervision sources}
For the Spouses dataset, and the IMDB variant with 12 LFs, we use the same LFs as in \cite{triplets} and \cite{TripletsMean}, respectively\footnote{All necessary label matrices are available in our research source code. The Spouse LFs and data are also available at the following URL:  \url{https://github.com/snorkel-team/snorkel-tutorials/blob/master/spouse/spouse_demo.ipynb}}.
The set of 12 IMDB LFs was specifically chosen to have a large coverage, see Table \ref{tab:datasets}.
These LFs and the larger set of LFs that we introduce for the second IMDB experiment are all pattern- and regex-based heuristics, i.e. LFs that label whenever a certain word or bi-gram appears in a text document. For instance, 'excellent' would label for the positive movie review sentiment (and would do so with $80\%$ accuracy on the samples where it does not abstain). This holds for the other text datasets as well, while the Spouse experiments also contain LFs that are distant supervision sources based on DBPedia. 
\\
For the remaining datasets (IMDB with 136 LFs, Bias Bios, and Amazon), we created the respective LF sets ourselves, prior to running experiments.

\paragraph{Encoder network architectures}
In all experiments, we use a simple multi-layer perceptron (MLP) as the encoder $e$, with two hidden layers, batch normalization, and ReLU activation functions. 
For the Spouse dataset, we use a bottleneck-structured network of sizes 50, 5. This is motivated by the small size of the set of samples labeled by at least one LF. For all other datasets we use hidden dimensions of 70, 70. We show in the ablations (Table \ref{tab:fullAbl}), that our end-to-end model also succeeds for different encoder architecture choices.
\paragraph{Downstream models}
For all datasets besides Spouse, we use a three-layer MLP with hidden dimensions 50, 50, 25.
For Spouse, we use a single-layer bidirectional LSTM with a hidden dimension of 150, followed by two fully-connected readout layers with dimensions 64, 32.
All fully-connected, layers use ReLU activation functions.
We choose simple downstream architectures as we are interested in the relative improvements over other label models. More sophisticated architectures are expected to further improve the performances, however. 
\paragraph{Hyperparameters}
Unless explicitly mentioned, all reported experiments are averaged out over seven random seeds.
We use an L2 weight decay of $7\text{e-}7$ and dropout of $0.3$ for both encoder and downstream model for all datasets but Spouse (where the LSTM does not use dropout).
All models are optimized with Adam, with early-stopping based on AUC performance on the small validation set, and a maximum number of $150$ epochs ($75$ for Spouse). The batch size is set to $64$.
The loss function is set to the (binary) cross-entropy.
For each dataset and each model/baseline, we run the same experiment for learning rates of $1\text{e-}4$ and $3\text{e-}5$, and then report the model chosen according to the best ROC-AUC performance on the small validation set. 
For Spouse we additionally run experiments with a L2 weight decay of $1\text{e-}4$ which due to the risk of overfitting to the small size of LF-covered data points boosts performance for all models. 
For our own model, \weasel, we also run additional experiments for Spouses with different configurations of the temperature hyperparameter, $\tau_1 \in \{1, 1/3\}$ and again report the test performance as measured by the best validation ROC-AUC. \\
The probabilistic labels from Snorkel used for downstream model training are chosen over six different configurations of the learning rate and number of epochs for Snorkel's label model (again with respect to validation set ROC-AUC).
For all binary classification datasets (i.e. all except for LabelMe), we tune the downstream model's decision threshold based on the resulting F1 validation score for all models. We believe that this, alternatively to reporting test ROC-AUC scores, makes the comparison fairer, since F1 is a threshold dependent metric.
All label model baselines are provided with the class balance, which \weasel\ does not use (but which is expected to be helpful for unbalanced classes, where no validation set is available).

\section{Extended Ablations} \label{sec:ablationsAppendix}
\begin{table}
\centering
\caption{Ablative study on the subcomponents of our algorithm as in Alg.~\ref{alg:main} (over 5 random seeds). In each row below we change exactly one component of \weasel\, and report the resulting F1 score. Note that the scores for \weasel\ are slightly different to the ones in the main results table, since they were run separately, with fewer seeds, and for only one learning rate (1e-4). Configurations that \textbf{outperform base \weasel\ are highlighted in bold font}, while the \color{red}\textbf{four worst performing configurations }\color{black} are highlighted in red for each dataset. Note that bold font does not indicate significant differences.}
\vspace{5pt}
\label{tab:fullAbl}
\begin{tabular}{@{} *5l @{}}
\toprule
    Change  & 
        ProfTeacher & IMDB-$136$ \small{LFs} & IMDB-$12$ \small{LFs} & Amazon \\ 
    \midrule
    \weasel & 
        $86.8 \pm 0.4$ & $82.1 \pm 0.7$ &  $77.3 \pm 0.5$ & $86.6 \pm 0.5$ \\ 
    \midrule
    $\theta(\lfs, \features) = \theta(\lfs)$ & 
        $85.6 \pm 1.6$ & $82.1 \pm 0.5$ & $75.9 \pm 0.8$ & \Abl{86.6}{0.4} \\
    Linear $e$ &
        $81.9 \pm 0.7$ & $80.0 \pm 0.6$ & $73.2 \pm 0.6$ &  $82.6 \pm 0.5$ \\
    1 hidden layer $e$ &
        \emphAbl{87.1}{0.7} & $81.8 \pm 0.6$ & $76.8 \pm 0.9$ & $85.3 \pm 0.8$\\
    75x50x25x50x75 $e$ &
        $84.3 \pm 2.1$ & $81.9 \pm 0.6$ & $75.8 \pm 1.1$ & $86.1 \pm 0.6$ \\
    $\tau_1 = 2$   & 
        $86.7 \pm 1.0$ & $81.9 \pm 0.3$ & $77.3 \pm 0.5$ & $85.5 \pm 1.0$ \\ 
    $\tau_1 = 1/2$   & 
        $86.5 \pm 0.8$ & $81.8 \pm 0.5$ & $76.0 \pm 1.4$ & $86.4 \pm 0.3$ \\ 
    $\tau_1 = 1/4$   & 
        $84.5 \pm 1.2$ & $81.8 \pm 0.2$ & $73.9 \pm 0.9$ &  $85.6 \pm 1.0$ \\
        
    $\tau_2 = 1$   & 
        $85.2 \pm 1.6$ & \emphAbl{82.2}{0.4} & $76.6 \pm 1.0$ & $84.3 \pm 1.2$\\
    $\tau_2 = m$   & 
        $86.1 \pm 0.7$ & $81.2 \pm 0.6$ & $76.4 \pm 0.4$ & $85.7 \pm 0.2$ \\
    No BatchNorm &
        $82.6 \pm 1.4$ & $81.9 \pm 0.5$ & $74.7 \pm 0.7$ & $85.3 \pm 0.8$ \\
    1e-4 weight decay &
        \emphAbl{87.4}{0.4} & $80.9 \pm 1.3$ & \emphAbl{77.9}{0.6} & $85.2 \pm 0.5$\\
    MIG loss &
        $86.7 \pm 0.4$ & $78.7 \pm 0.4$ & $74.1 \pm 0.4$ & $84.7 \pm 1.8$\\
    L1 loss &
        $86.2 \pm 0.6$ & $81.1 \pm 0.5$ & $75.6 \pm 0.9$ & $84.1 \pm 0.9$\\
    Squared Hellinger loss &
        \emphAbl{87.4}{0.3} & \emphAbl{82.2}{0.6} & $75.7 \pm 1.1$ & $86.3 \pm 0.4$ \\
    CE$\brackets{P_f, P_e}$ asymm. loss &
            \emphBadAbl{77.3}{3.7} & \emphBadAbl{77.7}{1.1} &  \Abl{71.7}{0.3} & \emphBadAbl{78.7}{1.2}\\
    CE$\brackets{P_e, P_f}$ asymm. loss &
            \emphBadAbl{73.1}{6.8} & \emphBadAbl{71.9}{1.9} & \emphBadAbl{69.7}{0.7} & \emphBadAbl{70.1}{1.1}\\
    No \texttt{stop-grad} &
      \emphBadAbl{80.4}{2.1} & \emphBadAbl{76.2}{0.5}  & \emphBadAbl{71.0}{0.6} & $79.3 \pm 0.6$\\
    $\theta(\lfs, \features) = \sqrt{m}\cdot\text{sigmoid}(e(\lfs, \features))$ &
        $85.5 \pm 0.6$ & $81.8 \pm 0.5$ & \emphAbl{78.0}{0.7} & \emphAbl{86.9}{0.3}\\
    $\theta(\lfs, \features) = \text{ReLU}(e(\lfs, \features)) +$ 1e-5 &
        $83.0 \pm 2.3$ & $78.3 \pm 1.1$ & \emphBadAbl{69.1}{2.1} & \emphBadAbl{74.2}{2.7}\\
    $\theta(\lfs, \features) = \text{Tanh}(e(\lfs, \features))$ &
        \emphBadAbl{71.9}{4.0} &  \emphBadAbl{67.0}{0.8} & \emphBadAbl{67.0}{1.1} &  \emphBadAbl{67.3}{1.1} \\
\bottomrule
\end{tabular}
\end{table}

\begin{figure}
    \begin{subfigure}{.99\textwidth}
      \centering
      \includegraphics[width=.99\linewidth]{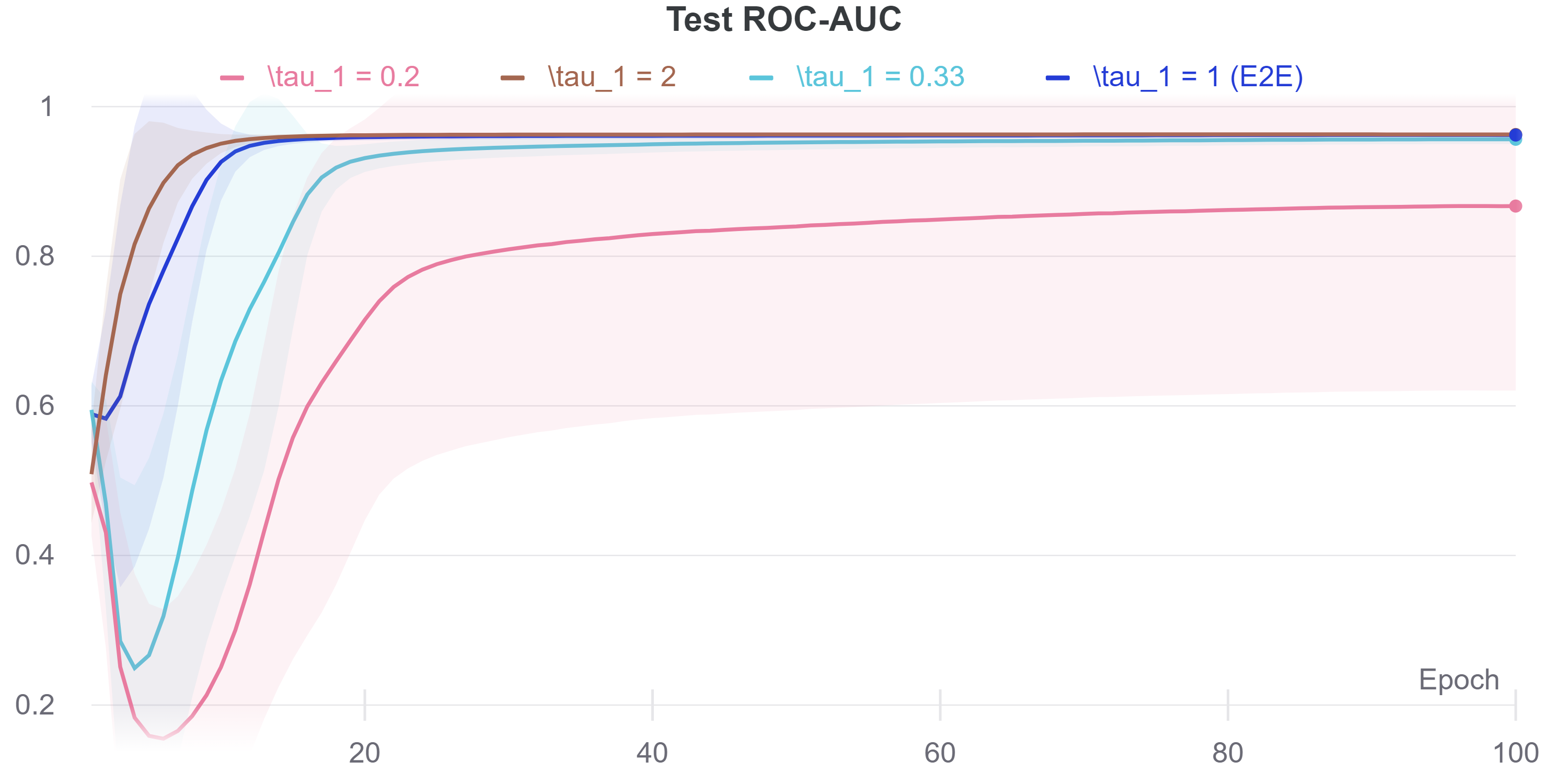}
      \label{fig:synthRobustnessTau1Log}
    \end{subfigure}%
    \caption{
    Test AUC performance at each training epoch for different choices of $\tau_1 \in \{1/5, 1/3, 1, 2\}$ on our synthetic experiment, see appendix \ref{sec:synthRobustnessDuplicatesAppendix}, averaged out over the number of duplicates and five random seeds. 
    A lower $\tau_1$ leads to slower or worse convergence in this specific case. A lower $\tau_1$ corresponds to smoother accuracies, which makes their induced label depend on more LFs. Since in this specific case only one LF is 100\% accurate and the rest are not better than a coin flip, the shown behavior is expected.
    }
  \label{fig:synthRobustnessTau1}
\end{figure}

The full ablations are reported in Table \ref{tab:fullAbl}, where in each row we change or remove exactly one component of our proposed model, \weasel.
We find that the design choices of \weasel\ which were inspired by sensible inductive biases for an encoder label model are hard to beat by various changes to the architecture, loss function, or hyperparameters.
Indeed, most changes consistently underperform \weasel, and the occasional positive changes -- 1e-4 weight decay, and the Squared Hellinger loss instead of the symmetric cross-entropy -- only beat the base \weasel\ performance in at most two datasets, and never significantly. In practice, we advise to explore these strongest configurations if a small validation set is available.
\\
We find that letting the accuracy scores depend on the input features (first row), usually boosts performance, but not by much (1.2 F1 points at most).
On the other hand, it proves very important to allow these accuracy scores to depend non-linearly on the LF votes and the features: A linear encoder network, as in \cite{MaxMIG}, significantly underperforms \weasel\ with at least one hidden layer by up to 4.9 F1 score points.
Conversely, a deeper encoder network (of hidden dimensionalities $75,50,25,50,75$, see fourth row) does not improve results. This may be due to the sample-dependent accuracies not being a too complex function to learn.  
\\
While the effect of the inverse temperature parameter $\tau_1$--which controls the softness of the encoder-predicted accuracy scores--on downstream performance is not large, it can have significant effects on the learning dynamics and robustness, see Fig \ref{fig:synthRobustnessTau1} for such learning curves as a function of epoch number. In particular, a lower $\tau_1$ makes the dynamics more robust, since the accuracy score weights are more evenly distributed across LFs, which appears to help avoid overfitting. When overfitting is not easily detectable due to a lack of a validation set, it is therefore advisable to use a lower $\tau_1$.
It also proves helpful to scale the $\softmax$ in Eq. \ref{eq:AccuracyScores} by $\sqrt{m}$, rather than not scaling it ($\tau_2 = 1$ row) or scaling by $m$.
\\
Changing the loss function from the symmetric cross-entropy to the MIG function \cite{MaxMIG} or the L1 loss consistently leads to worse performance. The former is interesting, since using the MIG loss for the crowdsourcing dataset LabelMe, see subsection \ref{sec:crowdsourcing}, was important in order to achieve state-of-the-art crowdsourcing performance (with a similar lift in performance observable for Snorkel using MIG for downstream model training).
The result provides some evidence that the MIG loss may be inappropiate for weak supervision settings other than crowdsourcing, while its use may be recommended for that specific setting.
\\
We find that it is important to constrain the accuracy score space to a positive interval, either by viewing them as an aggregation of the LFs via the scaled $\softmax$ in Eq. \ref{eq:AccuracyScores}, or by replacing the $\softmax$ with a sigmoid function. Indeed, using a less constrained activation function for the estimated accuracies (last two rows, where the 1e-5 in the ReLU row avoids accuracy scores equal to zero) significantly underperforms:
Allowing the accuracies to be negative (last row) leads to collapse and bad downstream performance. This is likely due to the removal of the inductive bias that LFs are better-than-random, which makes the joint optimization more likely to find trivial solutions.
Additionally, we find that our choice of using the symmetric cross-entropy loss with \texttt{stop-grad} applied to the targets is  crucial for the strong performance of \weasel. Removing the \texttt{stop-grad} operation, or using the standard cross-entropy (without \texttt{stop-grad} on the target) leads to significantly worse scores and a very brittle model. This is somewhat expected, since conceptually our goal is to have an objective that maximizes the agreement between a pair of models that predict based on two different views of the latent label, the features and the LF votes. The cross-entropy with \texttt{stop-grad} on the target\footnote{or, due to the \texttt{stop-grad} operation, equivalently the KL divergence} naturally encodes this understanding, since each model uses the other model's predictions as a reference distribution. 
Losses that already are symmetric (e.g. L1 or Squared Hellinger loss) neither need to be symmetrized nor use \texttt{stop-grad}. While the L1 loss consistently underperforms, we find that the Squared Hellinger loss can lead to better performance on two out of four datasets.
\\
However, only the symmetric cross-entropy loss with \texttt{stop-grad} on the targets is shown to be robust and able to recover the true labels in our synthetic experiments in appendix \ref{sec:robustnessAppendix}, see Fig. \ref{fig:syntheticRobustnessAppendix} in particular. 
The synthetic ablation in appendix \ref{sec:robustnessAppendix} gives interesting insights, and strongly supports the proposed design of \weasel. Indeed, many choices for \weasel\ that perform well enough on the real datasets, such as no features for the encoder, $\tau_2 = 1$, sigmoid parameterized accuracies, and all other objectives that we evaluated, lead to significantly worse performance and less robust learning on the synthetic adversarial setups.

\section{Crowdsourcing dataset} \label{sec:crowdsourcingAppendix}
As the crowdsourcing dataset, we choose the multi-class LabelMe image classification dataset that was previously used in the most related crowdsourcing literature \cite{rodrigues2018deep, MaxMIG}.
Note that this dataset consists of $10k$ samples, of which only $1k$ are unique, in the sense that the rest are augmented versions of the $1k$. They were annotated by $59$ crowdworkers, with a mean overlap of $2.55$ annotations per image.
The downstream model is identical to the previously reported one \cite{rodrigues2018deep, MaxMIG}. That is, a VGG-16 neural network is used as feature extractor, and a single fully-connected layer (with 128 units and ReLU activation) and
one output layer is put on top, using 50 \% dropout.

Experiments were conducted over seven random seeds with a learning rate of 1e-4 and 50 epochs. The reported scores are the ones with best validation set accuracy for a L2 weight decay $\in \{$ 7e-7, 1e-4 $\}$. The validation set is of size 200, and was split at random from the training set prior to running the experiments.
\\
As is usual in the related work for multi-class settings~\cite{Snorkel}, we employ class-conditional accuracies $\theta(\lfs, \features) \in \R^{m \times C}$ instead of only $m$ class-independent accuracies. Recall the LF outputs indicator matrix,
$\lambdaOnehot \in \R^{m \times C}$. To compute the resulting output softmax logits $\mathbf{s} \in \R^{C}$, we set $\mathbf{A} = \theta(\lfs, \features) \odot \lambdaOnehot \in \R^{m \times C}$ and $\mathbf{s}_j = \sum_i \mathbf{A}_{ij} \in \R$, where $\odot$ is the element-wise matrix product and we sum up the resulting matrix $\mathbf{A}$ across the LF votes dimension.
\\
Snorkel+MIG indicates that the downstream model $f$ was trained on the MIG loss with respect to soft labels generated by the first Snorkel step, label modeling. Snorkel+CE refers analogously to the same training setup, but using the cross-entropy (CE) loss.
All crowdsourcing baseline models are based on the open-source code from \cite{MaxMIG}.

\section{Robustness experiments}
\label{sec:robustnessAppendix}
In this section we give more details on the experiments that validate the robustness of our approach against (strongly) correlated LFs that are not better than a random coin flip. In addition, we present one further experiment where the random LFs are independent of each other -- a more difficult setup for learning (but which does not violate any assumptions of the PGM-based methods) -- and our model, \weasel, again is shown to be robust to a large extent.
\\
In contrast to \weasel, prior PGM-based work \cite{Snorkel, triplets, TripletsMean} attain significantly worse performance under these settings, due to assuming a Naive Bayes generative model where the weak label sources are conditionally independent given the latent label.
\subsection{Adversarial LF duplication}
\label{sec:advRobustnessAppendix}
For this experiment we use our set of 12 LFs for the IMDB dataset and generate a fake adversarial source by flipping the abstain votes, of the 80\%-accurate LF that labels for the positive sentiment on 'excellent', to negative ones.
\subsection{Recovery of true labels under massive LF noise}
\label{sec:synthRobustnessAppendix}
In this set of synthetic experiments we again validate the robustness of our approach. We focus on the Bias in Bios dataset, and use the features and true labels, $y^*$, therein. We let our initial LF set consist of 1) a 100\% accurate LF, that is we set $\lambda_1 = y^*$, and 2) a LF that votes according to the class balance (i.e. a coin flip with probabilities for tail/head set according to the class balance), i.e. $\lambda_2 \sim P(y)$.
In the first experiment we then add the same random LF $\lambda_2$ multiple times into the LF set (i.e. we duplicate it), see \ref{sec:synthRobustnessDuplicatesAppendix}, while in the second one, we incrementally add random LFs independently of $\lambda_2$ (and independently of any other LF already in the LF set), see \ref{sec:synthRobustnessIndependentAppendix}.
For both setups, our model, \weasel, is able to recover the performance of the same downstream model, $f$, that is directly trained on the true labels, $y^*$ (F1 $= 90.65$, ROC-AUC $=0.967$, see Table \ref{tab:tableResultsAppendix}).
In contrast, the PGM-based baselines quickly collapse.

\subsubsection{Random LF duplication}
\label{sec:synthRobustnessDuplicatesAppendix}
\begin{figure}[htb]
\centering
 \begin{subfigure}[b]{.48\linewidth}
    \centering
    \includegraphics[width=.99\textwidth]{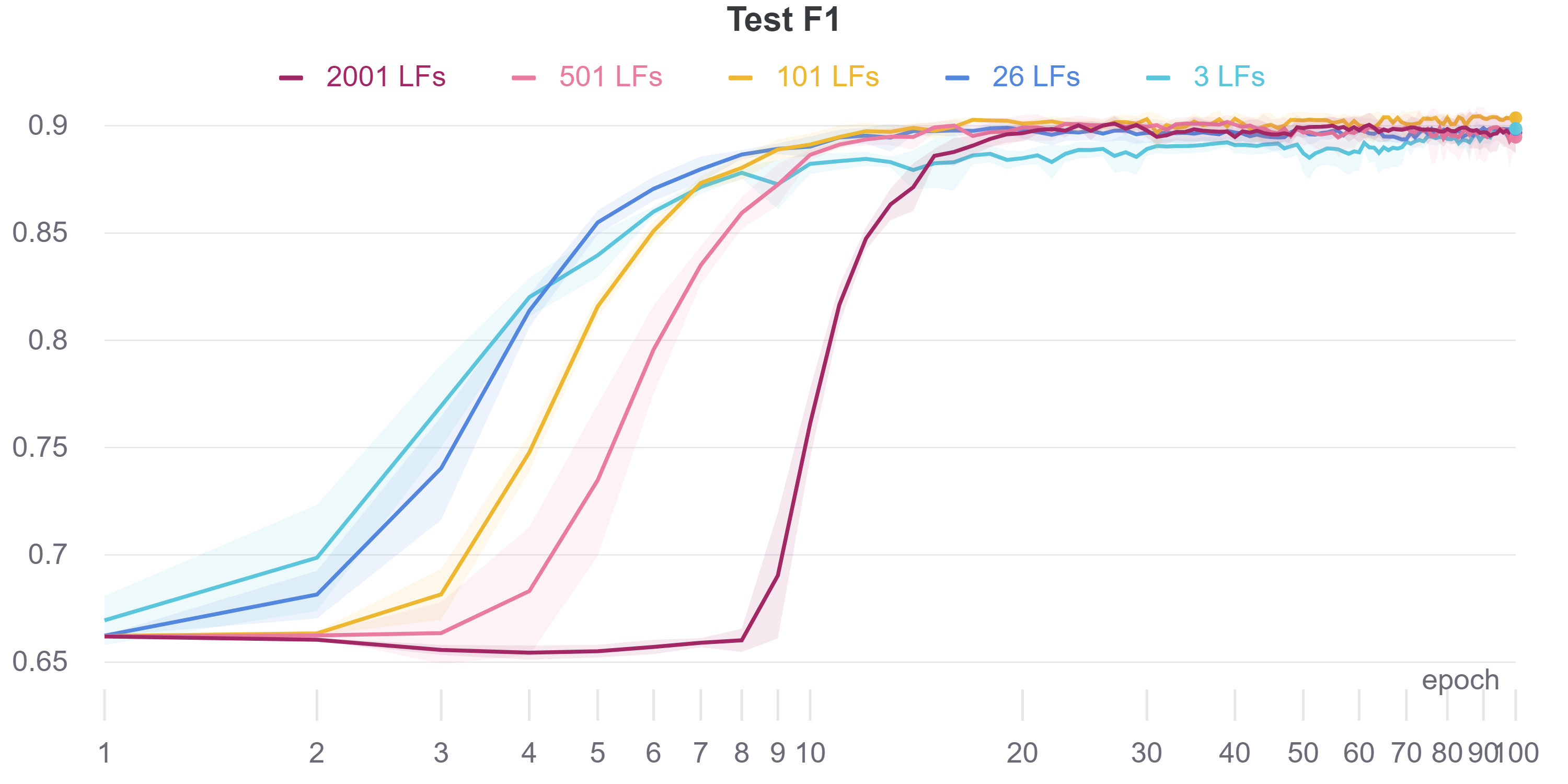}
    \caption{\weasel\ log-scale F1}\label{fig:logWeaselF1}
    \end{subfigure}%
  \begin{subfigure}[b]{.48\linewidth}
    \centering
    \includegraphics[width=.99\textwidth]{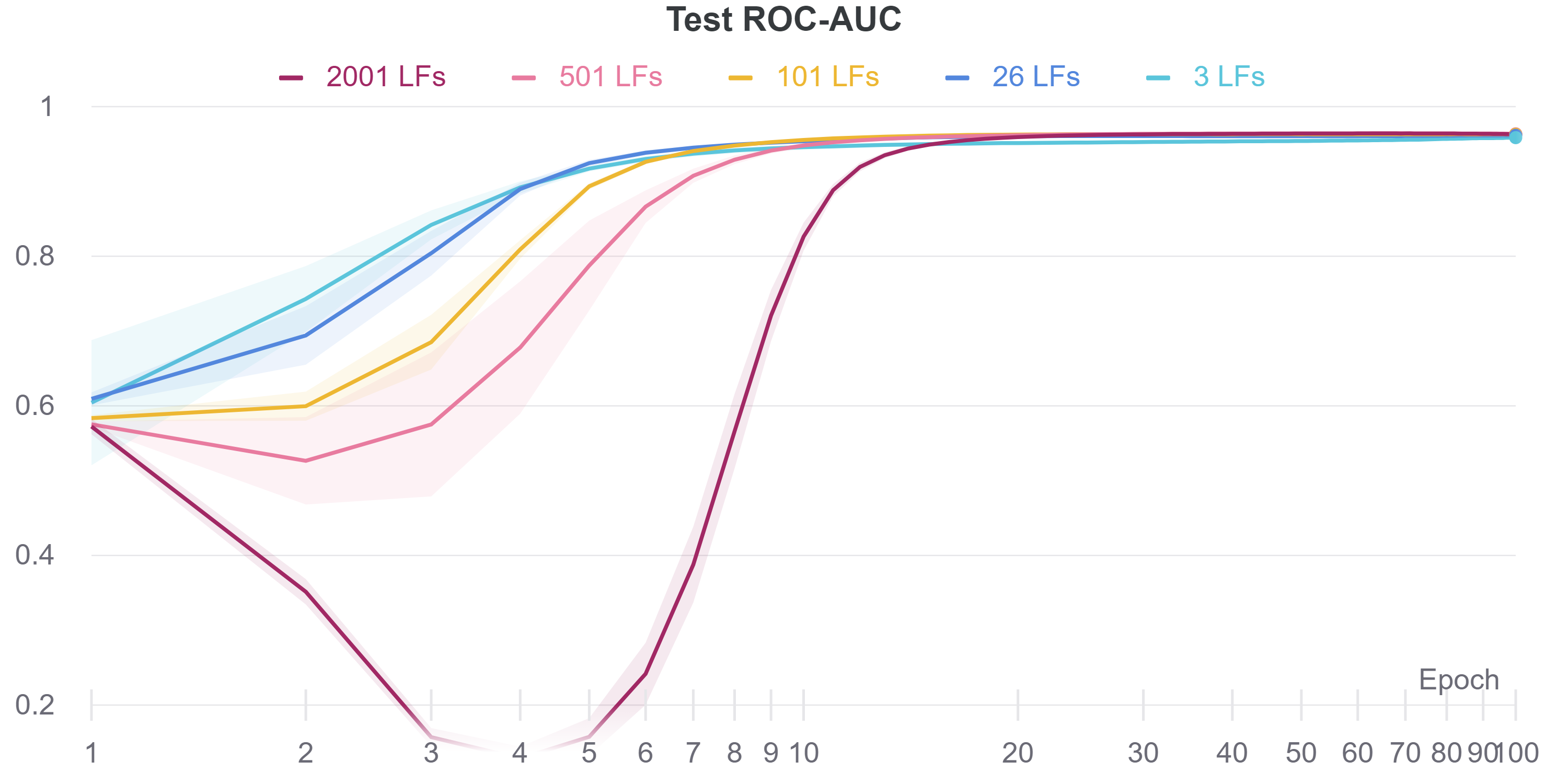}
    \caption{\weasel\ log-scale AUC}\label{fig:logWeaselAUC}
 \end{subfigure} \\
  \begin{subfigure}[b]{.48\linewidth}
    \centering
    \includegraphics[width=.99\textwidth]{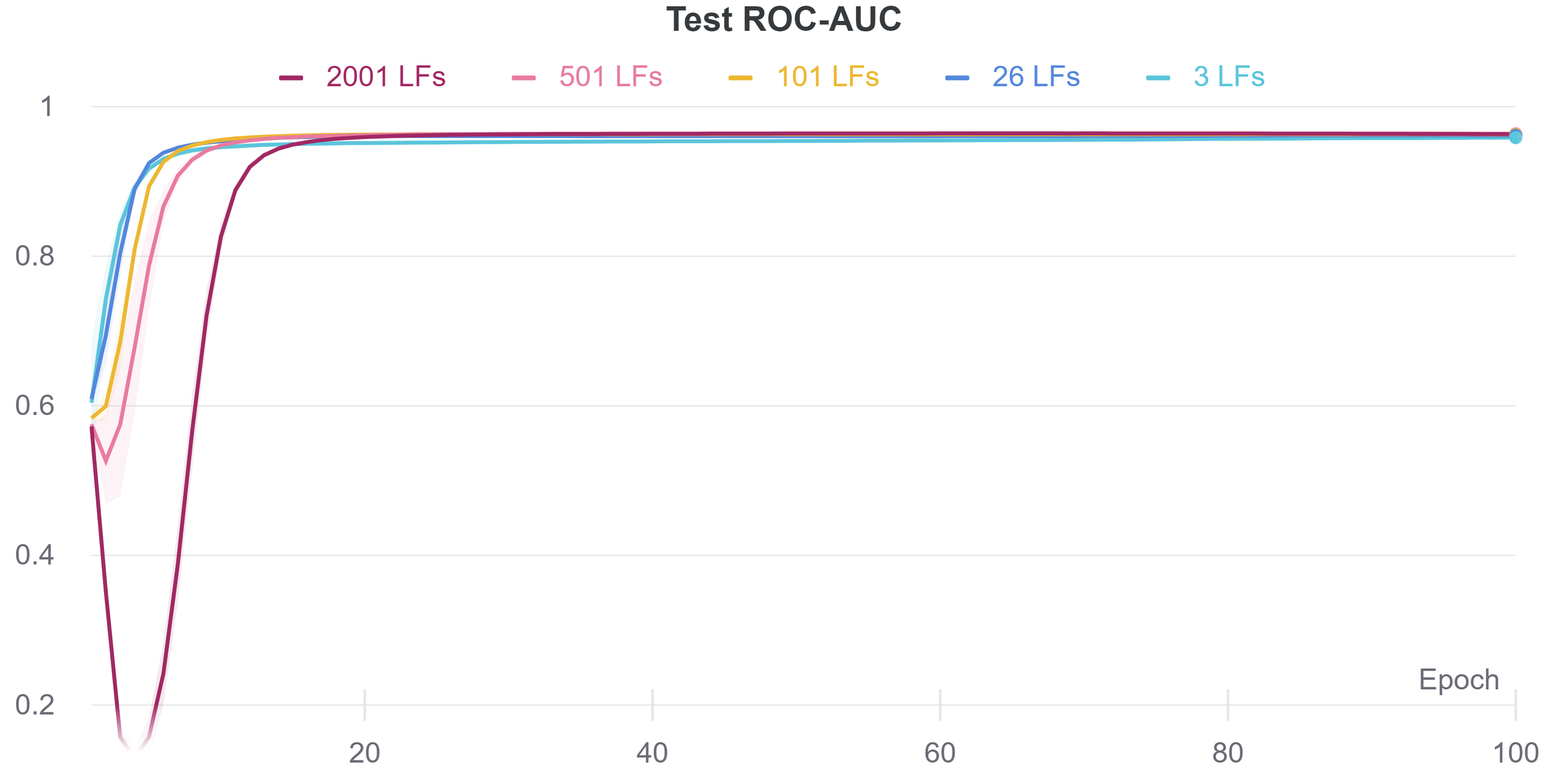}
    \caption{\weasel}\label{fig:WeaselAUC}
  \end{subfigure}%
  \begin{subfigure}[b]{.48\linewidth}
    \centering
    \includegraphics[width=.99\textwidth]{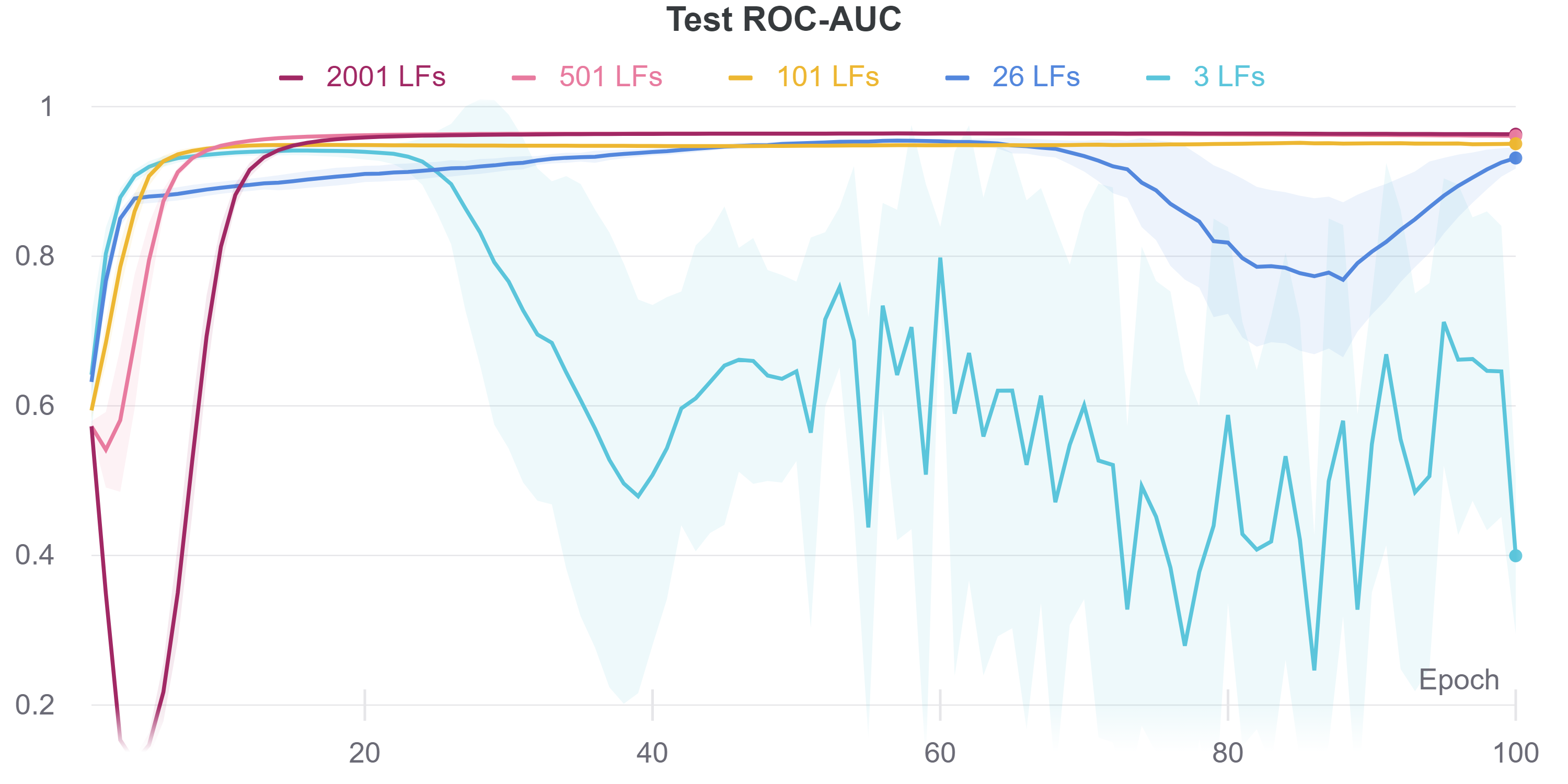}
    \caption{No features for encoder}\label{fig:noFeats}
 \end{subfigure} \\
  \begin{subfigure}[b]{.48\linewidth}
    \centering
    \includegraphics[width=.99\textwidth]{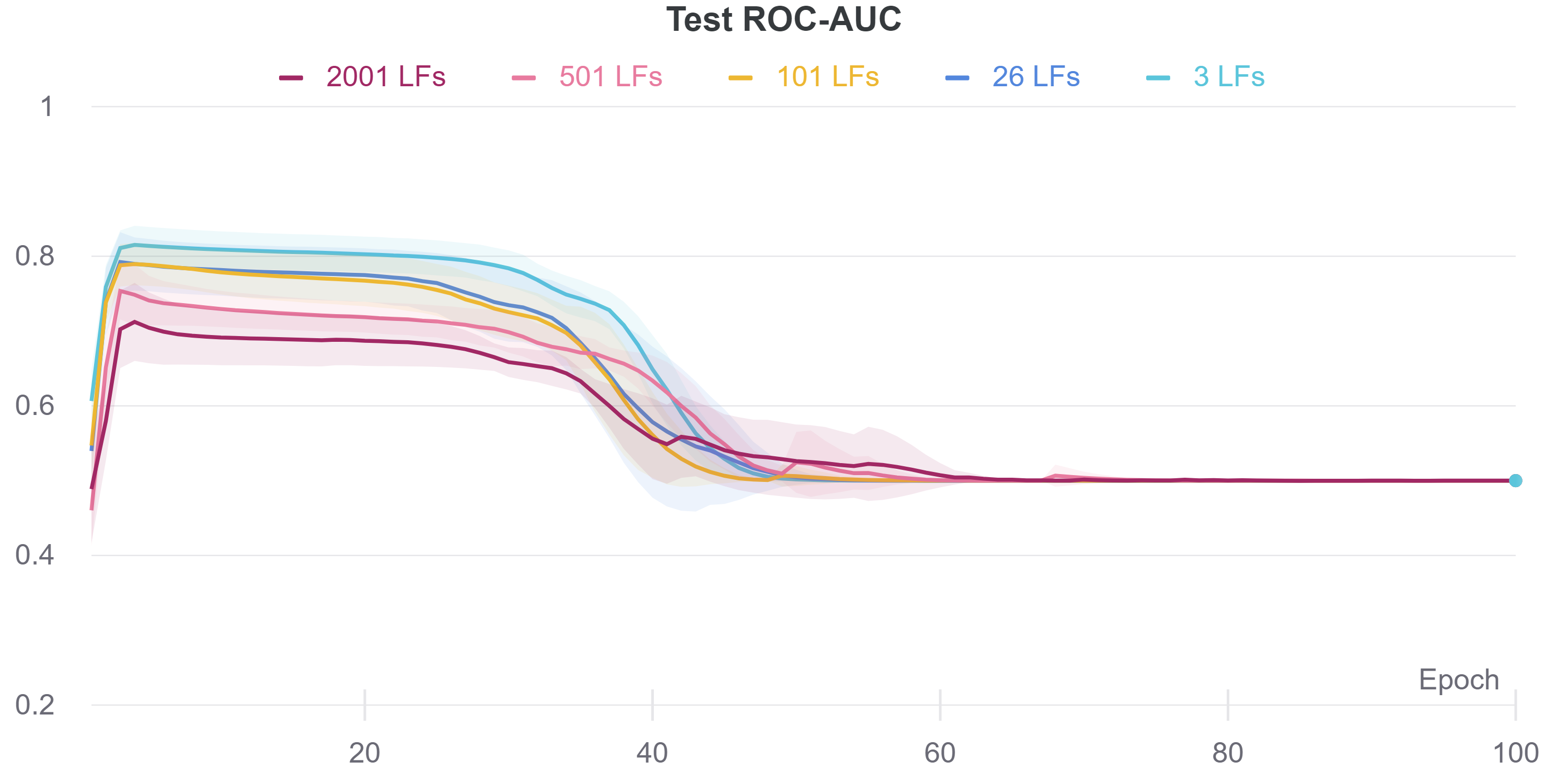}
    \caption{No \texttt{stop-grad}}\label{fig:noStopGrad}
  \end{subfigure}%
  \begin{subfigure}[b]{.48\linewidth}
    \centering
    \includegraphics[width=.99\textwidth]{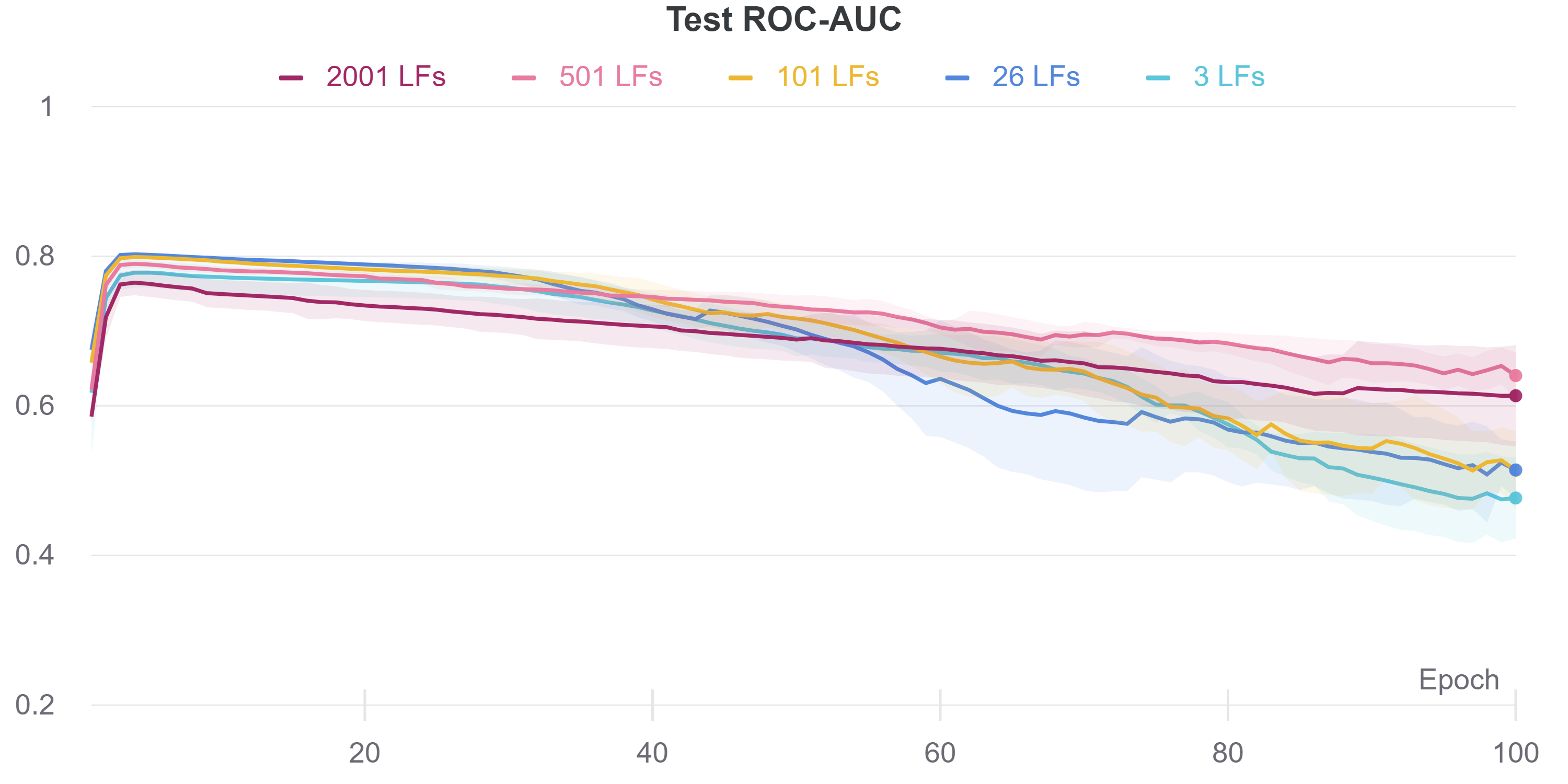}
    \caption{Asymmetric CE}\label{fig:asymCE}
 \end{subfigure} \\
  \begin{subfigure}[b]{.48\linewidth}
    \centering
    \includegraphics[width=.99\textwidth]{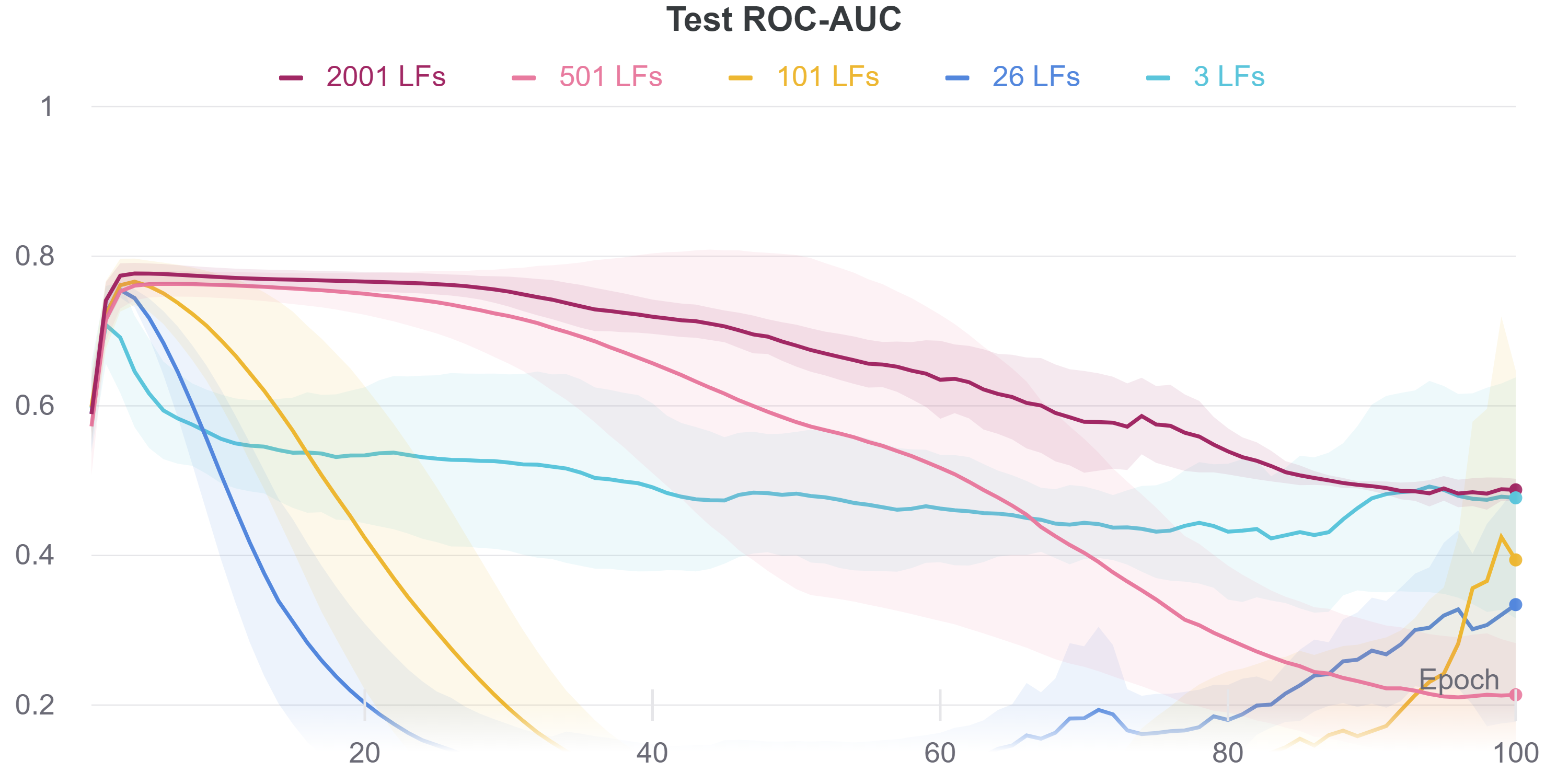}
    \caption{Sigmoid accuracies}\label{fig:SigmoidAccs}
  \end{subfigure}%
  \begin{subfigure}[b]{.48\linewidth}
    \centering
    \includegraphics[width=.99\textwidth]{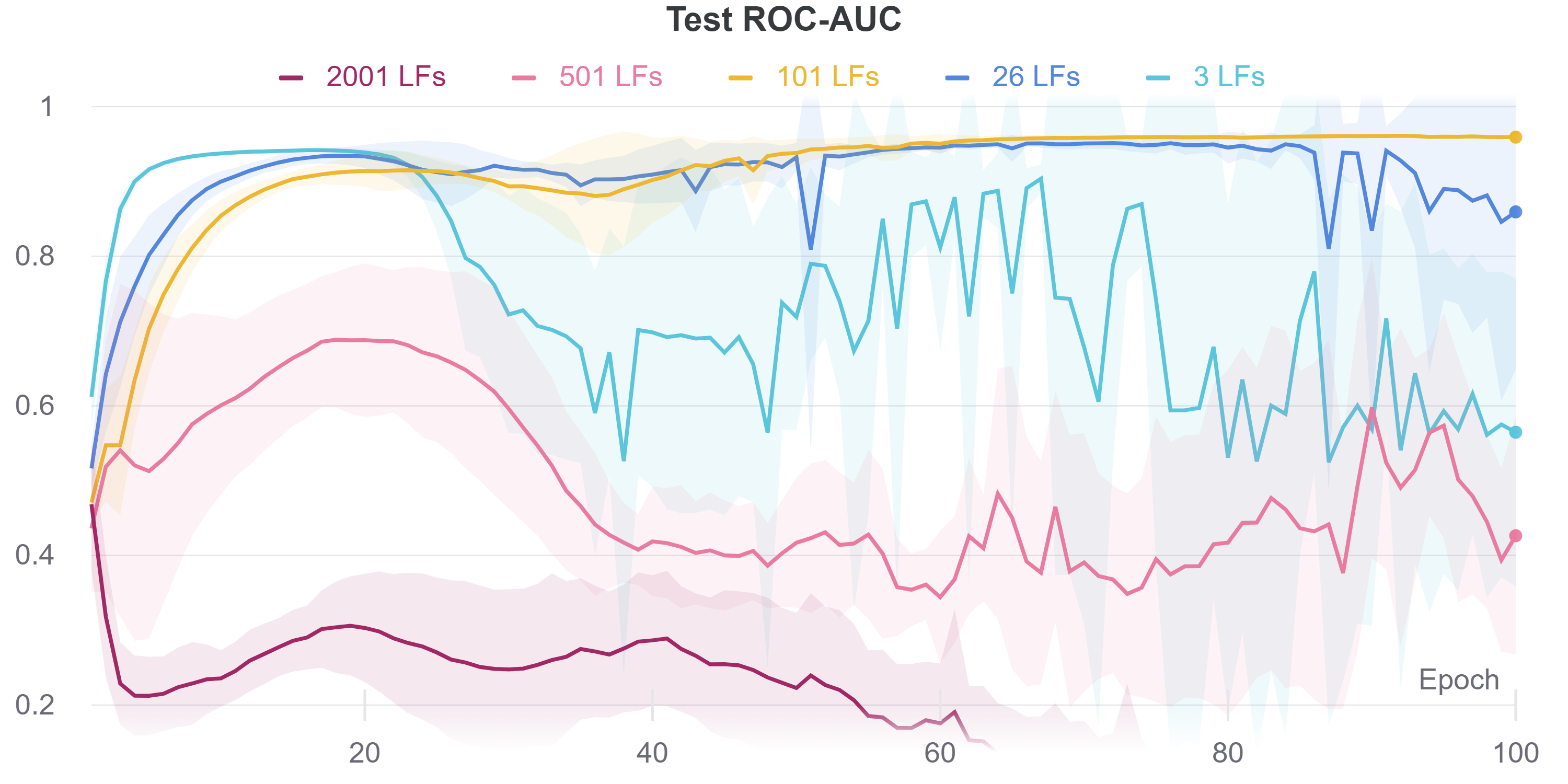}
    \caption{$\tau_2 = 1$}\label{fig:noScaler}
 \end{subfigure} \\
  \caption{We start with a 100\% accurate LF (i.e. ground truth labels) and plot test performances at each training epoch for a varying number of duplicates $\in \{2, 25, 100, 500, 2000\}$ of a LF that is no better than a coin flip. Performances are averaged out over five random seeds, and the standard deviation is shaded. More details are given in \ref{sec:synthRobustnessDuplicatesAppendix}.}
  \label{fig:syntheticRobustnessAppendix}
\end{figure}

This experiment is inspired by the theoretical comparison in Appendix E of \cite{MaxMIG} between the authors' end-to-end system and maximum likelihood estimation (MLE) approaches that assume mutually independent LFs. 
The authors show that such MLE methods are not robust against the following simple example with correlated LFs.
Based on the setup described above in \ref{sec:synthRobustnessAppendix}, we duplicate the random LF $\lambda_2$ multiple times, i.e. $\lambda_3 = \dots = \lambda_m=\lambda_2$. 
We run experiments for varying number of duplicates $\in \{2, 25, 100, 500, 2000\}$.
With this synthetic set of $m$ LFs, where one LF is $100\%$ accurate while the other $m-1$ LFs are just as good as a random guess, we train \weasel\ in the usual way on the features from the Bias in Bios dataset as well as the corresponding, just created, LF votes.
\\
\weasel\ is able to consistently and almost completely \emph{recover this fully supervised performance, even when the number of duplicates is very high} ($m=2001$). 
Snorkel and triplets methods, on the other hand, fare far worse (AUC $\approx 0.5$) for all numbers of duplicates. This behavior is similar to the one observed in \ref{sec:advRobustnessAppendix} (see Fig. \ref{fig:mainRobustness} for the performance of the baselines and \weasel\ averaged out over the varying number of duplicates, and Fig. \ref{fig:syntheticRobustnessAppendix}a-c for the separate performance of \weasel\ for each number of duplicates).

We also run an additional ablation study on this synthetic experiment that shows that the observed robustness does not hold for all configurations of \weasel. 
In Fig. \ref{fig:syntheticRobustnessAppendix} we plot the test performance curves over the training epochs for each number of LF duplications.
\\
Our proposed model, \weasel\ enjoys a stable and robust test curve (Fig. \ref{fig:WeaselAUC}) and quickly recovers the fully supervised performance, even with 2000 LF duplicates (although convergence becomes slower as the LF set contains more duplicates).
On the other hand, we find that many other configurations and designs of \weasel\ lead to less robust and worse converging curves, collapses or bad performances.
Indeed, for this experiment it is key to use as the loss function the proposed symmetric cross-entropy with \texttt{stop-grad} applied to the targets (see Fig. \ref{fig:noStopGrad}, \ref{fig:asymCE}), accuracies parameterized by a scaled (Fig. \ref{fig:noScaler}) softmax (Fig. \ref{fig:SigmoidAccs}), and, to a lesser extent, using the features an input to the encoder (Fig. \ref{fig:noFeats}).
\\
While the impact of not using \texttt{stop-grad}, or using an asymmetric cross-entropy loss is similarly bad in the main ablations on our real datasets, other configurations, and in particular sigmoid-parameterized accuracies (the choice in \cite{astra}), an unscaled softmax, and no features for the encoder, often perform well there.
This additional ablation, however, provides support for why the good performances on the real datasets notwithstanding, our proposed design choices are most appropriate in order to attain strong test performances as well as stable and robust learning.

\subsubsection{Random, independent LFs}
\label{sec:synthRobustnessIndependentAppendix}
\begin{figure}
    \begin{subfigure}{.95\textwidth}
      \centering
      \includegraphics[width=.95\linewidth]{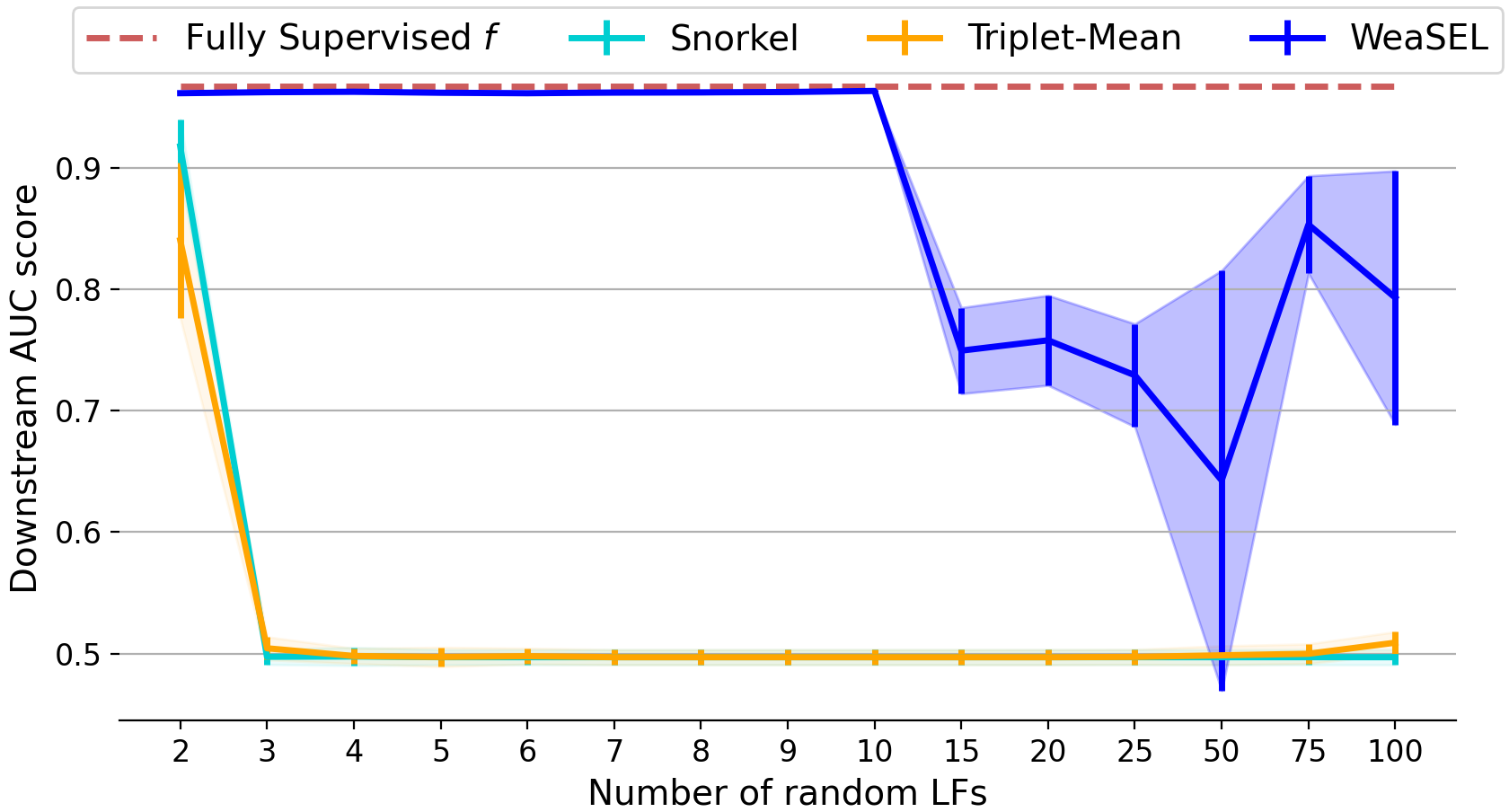}
      \label{fig:synthRobustnessIndepRandomLFsAUC}
    \end{subfigure}%
  \caption{We start with a 100\% accurate LF (i.e. ground truth labels) and incrementally add new, independent LFs that are no better than a random guess.
  \weasel\ recovers the performance of training directly on the ground truth labels (Fully Supervised $f$), for up to 10 such randomly voting LFs that are independent of each other.
  The PGM-based prior work, rapidly degrades in performance (AUC $\approx0.5$) and is not able to recover any of the $100$\% accurate signal of the true-labels-LF, as soon as the LF set is corrupted by three or more random LFs.
  Performances are averaged out over five random seeds, and the standard deviation is shaded. For more details, see \ref{sec:synthRobustnessIndependentAppendix}}
  \label{fig:synthRobustnessIndepRandomLFs}
\end{figure}
We start with the same setup as above in \ref{sec:synthRobustnessAppendix}, but instead of duplicating the same LF multiple times as in \ref{sec:synthRobustnessDuplicatesAppendix}, we now draw a new, independent random LF at each iteration.
That is, we start with  $\lambda_1 = y^*, \lambda_2 \sim P(y)$ as our initial LFs, and the incrementally add new LFs $\lambda_i \sim P(y)$ that have no better skill than a coin flip.
Note that this is arguably a harder setup than the one in the previous experiments, since there the LF set was corrupted by a single LF voting pattern. In this experiment, multiple equally bad, but independent, LFs corrupt the 100\% accurate signal of $\lambda_1$. Notably, since these $\lambda_2, \dots, \lambda_m$ are independent, we are not violating the independence assumptions of PGM-based methods.
Nonetheless, we find that these PGM-based baselines break with only three ($m=4$) of such random, but independent LFs, while \weasel\
is shown to be fully robust and able to recover the ground truth LF $\lambda_1$ for up to 10 random LFs ($m=11$). For more LFs, \weasel\ starts deteriorating in performance, but is still able to consistently outperform the trivial solution of voting randomly according to the class balance (i.e. based on $\lambda_2, \dots, \lambda_m$) and the baselines, see Fig. \ref{fig:synthRobustnessIndepRandomLFs}.

\section{Broader Impact} 
\label{sec:broaderImpact}
Large labeled datasets are important to many machine learning applications. 
Reducing the expensive human effort required to annotate such datasets is an important step towards making machine learning more accessible, more manageable, more beneficial, and therefore used more broadly. 
Our proposed end-to-end learning for weak supervision approach provides another step towards the practical utility of learning from multiple sources of weak labels on large datasets.
%
Methods such as the one presented in our paper must be applied with care. 
One of the risks to consider and mitigate in a particular application is the possibility of incorporating biases from subjective humans who chose weak labeling sources. This is particularly the case when heuristics might apply differently to different subgroups in data, such as may be the case in scenarios highlighted in recent research towards fairness in machine learning.

\end{document}